\documentclass[journal=jacsat,manuscript=article]{achemso}
\usepackage{array}
\newcolumntype{L}[1]{>{\raggedright\arraybackslash}p{#1}}

\usepackage{hyperref}
\usepackage{tabularx}
\usepackage{array}
\usepackage{makecell}
\usepackage{graphicx}
\usepackage{subcaption}
\usepackage{float}
\usepackage{stfloats}

\usepackage{siunitx}
\usepackage{bbm}
\usepackage{amsfonts}
\usepackage{mciteplus}
\mciteErrorOnUnknownfalse
\usepackage{listings}
\usepackage{xcolor}
\usepackage{longtable}
\usepackage{multirow}
\usepackage{amsmath}

\definecolor{jsonbg}{RGB}{245,245,245}
\definecolor{bluebg}{RGB}{230,240,255}
\definecolor{greenbg}{RGB}{230,255,230}

\definecolor{jsonbg}{gray}{0.97}
\lstdefinestyle{json}{
  basicstyle=\ttfamily\scriptsize,
  backgroundcolor=\color{jsonbg},
  frame=single,
  rulecolor=\color{gray!60},
  breaklines=true,
  showstringspaces=false,
  columns=fullflexible,
  keepspaces=true,
  tabsize=2,
  xleftmargin=1em,
  xrightmargin=1em,
  aboveskip=0.8em,
  belowskip=0.8em,
}

\lstdefinestyle{bluecode}{
    backgroundcolor=\color{bluebg},
    basicstyle=\ttfamily\small,
    frame=single,
    breaklines=true,
  showstringspaces=false,
  columns=fullflexible,
  keepspaces=true,
  tabsize=2,
  xleftmargin=1em,
  xrightmargin=1em,
  aboveskip=0.8em,
  belowskip=0.8em,
}

\lstdefinestyle{greencode}{
    backgroundcolor=\color{greenbg},
    basicstyle=\ttfamily\small,
    frame=single,
    breaklines=true,
  showstringspaces=false,
  columns=fullflexible,
  keepspaces=true,
  tabsize=2,
  xleftmargin=1em,
  xrightmargin=1em,
  aboveskip=0.8em,
  belowskip=0.8em,
}

\SectionNumbersOn

\usepackage[version=3]{mhchem} 
\usepackage{booktabs}
\usepackage{pifont} 
\newcommand{\cmark}{\ding{51}} 

\author{Achuth Chandrasekhar}
\affiliation[CMU-MECHE]
{Mechanical Engineering, Carnegie Mellon University, Pittsburgh, PA 15213, USA}

\author{Janghoon Ock}
\affiliation[unl-cheme]
{Department of Chemical and Biomolecular Engineering, University of Nebraska--Lincoln, Lincoln, NE 68588, USA}
\author{Amir Barati Farimani}
\email{barati@cmu.edu}
\affiliation[CMU-MECHE]
{Mechanical Engineering, Carnegie Mellon University, Pittsburgh, PA 15213, USA}
\alsoaffiliation[CMU-BME]
{Biomedical Engineering, Carnegie Mellon University, Pittsburgh, PA 15213, USA}
\alsoaffiliation[CMU-CHE]
{Chemical Engineering, Carnegie Mellon University, Pittsburgh, PA 15213, USA}
\alsoaffiliation[CMU-MSE]
{Materials Science and Engineering, Carnegie Mellon University, Pittsburgh, PA 15213, USA}
\alsoaffiliation[CMU-ML]
{Machine Learning Department, Carnegie Mellon University, Pittsburgh, PA 15213, USA}

\title[An \textsf{achemso} demo]
  {Catalyst-Agent: Autonomous heterogeneous catalyst screening with an LLM Agent}

\abbreviations{IR,NMR,UV}

\begin{document}

\begin{tocentry}
\centering
\includegraphics[width=0.8\textwidth]{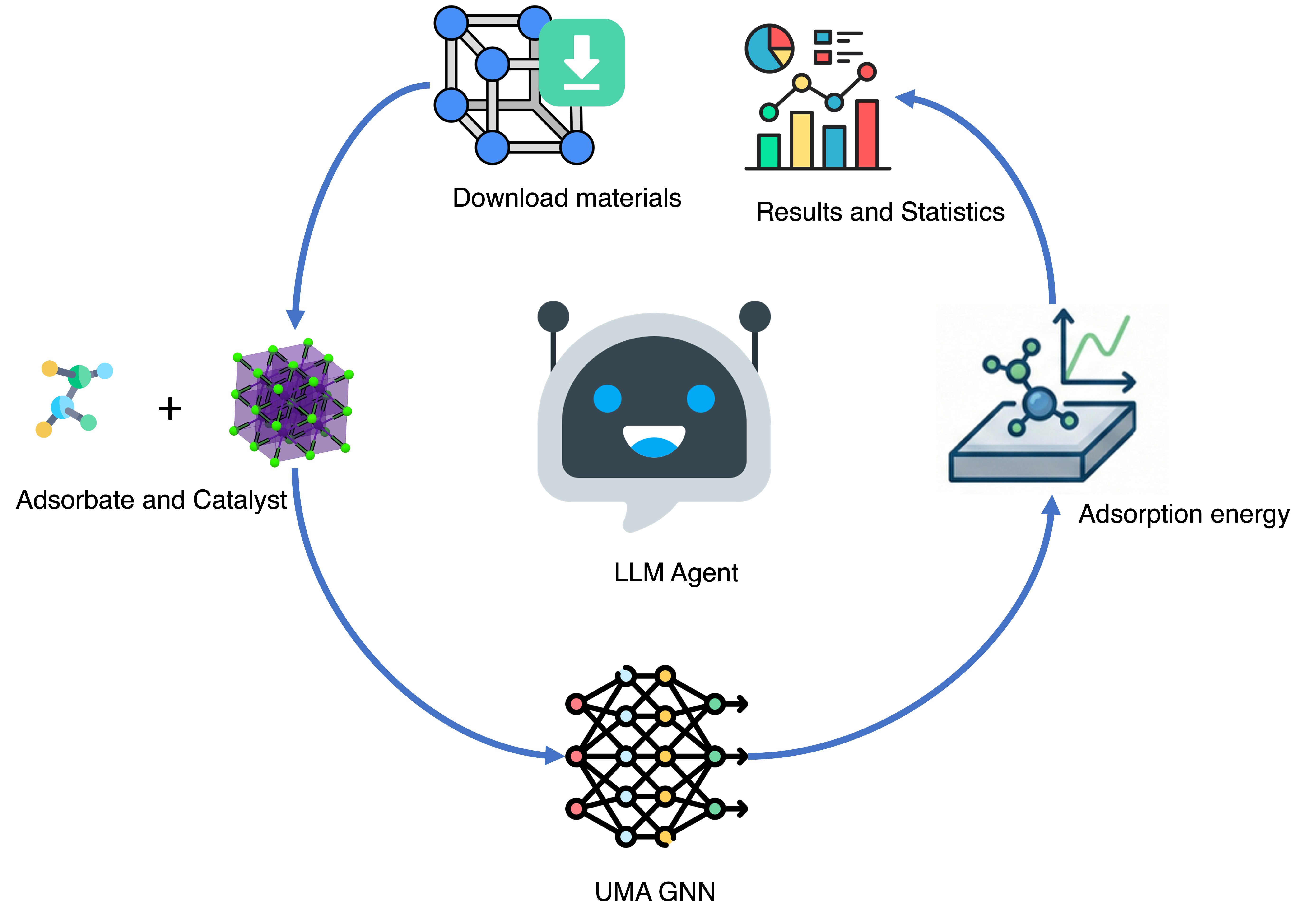}
\end{tocentry}

\begin{abstract}

The discovery of catalysts for electrochemical applications such as the oxygen reduction reaction (ORR), nitrogen reduction reaction (NRR), and CO$_2$ reduction reaction (CO$_2$RR) remains a central challenge in chemistry and materials science. Machine-learning interatomic potentials (MLIPs) and graph neural network models now accelerate individual adsorption-energy calculations by orders of magnitude relative to density functional theory. However, true large-scale screening is still blocked by human decisions: selecting candidates, constructing slabs, enumerating adsorption sites, interpreting descriptor failures, and choosing follow-up modifications. Here, we introduce Catalyst-Agent, a Model Context Protocol (MCP) server-based, LLM-powered agent that autonomously coordinates closed-loop catalyst screening. Catalyst-Agent searches materials databases through OPTIMADE, constructs slabs, computes adsorption energies using Meta FAIRchem's UMA MLIP within AdsorbML, evaluates reaction-specific descriptors, and applies structural modifications to refine near-miss candidates. In ORR, NRR, and CO$_2$RR campaigns, Catalyst-Agent demonstrates high performance and converges in 1.40--3.41 trials per successful material on average. It identified Sn$_3$Sc, Sn$_3$Y, Tl$_3$La, Pb$_3$Y and In$_3$Y as CO$_2$RR candidates for further validation that were not previously reported in the literature. DFT single-point checks confirmed screening outcomes for representative NRR and CO$_2$RR candidates. Ablations show these gains arise from chemically informed candidate selection and feedback-directed modification rather than brute-force evaluation: fully randomized screening dropped to 13.3\%, 16.7\%, and 0\% success for ORR, NRR, and CO$_2$RR, respectively. These results show that tool-grounded LLM agents can shift catalyst screening from manual trial-and-error toward more autonomous, reproducible and adaptive workflows.

\end{abstract}

\section*{Introduction}
\label{sec:introduction}
Designing heterogeneous catalyst materials that achieve high activity, selectivity, and stability for target chemical reactions is a central challenge in chemical technology to improve process efficiency and chemical sustainability across industries such as fertilizers, plastics and clean fuels.\cite{norskov2014fundamental, dumesic1996principles, zitnick2020introduction, BELRHAZI2026116098} Catalyst discovery has traditionally proceeded through iterative experimental cycles of hypothesis generation, synthesis, and performance measurement, with substantial time and monetary expenditure. In a sequential computational-to-experimental workflow, computational researchers first use first-principles electronic-structure calculations to predict catalytic trends and prioritize promising materials, after which experimental researchers synthesize the selected candidates and characterize their activity, selectivity, and stability. Density functional theory (DFT) enables estimation of adsorption energies, activation barriers and other energetic descriptors on well-defined catalytic surfaces, supporting scaling relations and mechanistic insights.\cite{raccuglia2016machine, wang2025density}
But catalyst screening remains bottlenecked not only by the complex search process over diverse materials, facets, adsorption sites and adsorbates, but also by the sequence of iterative decisions needed to evaluate hundreds of promising candidates at scale. At each stage, a human expert must choose candidate materials, decide which surfaces and adsorbates to evaluate, determine and interpret descriptors, and select the next modification or follow-up calculation. An autonomous agent that records explicit, tool-grounded reasoning traces can act as a proxy researcher for this decision-making process, turning the complex screening workflow into a reproducible chain of candidate selection, calculation, interpretation, and refinement.

Because exhaustive evaluation of every elementary step is impractical across large catalyst search spaces, descriptor-based screening is necessary to rapidly prioritize candidates with favorable reactivity. In heterogeneous catalysis, adsorption energies provide especially useful reactivity descriptors because they quantify how strongly key intermediates bind to catalyst surfaces and directly correlate with activity.\cite{norskov2009towards, yang2014understanding}For a given adsorbate, the relevant adsorption energy is defined as the global minimum energy over all candidate configurations:\cite{ock2023beyond, chen2025descriptor, karimadom2021calculating}

\begin{equation}
\Delta E_{\mathrm{i}} = E_{\mathrm{slab}+\mathrm{i}} - E_{\mathrm{slab}} - E_{\mathrm{gas}},
\end{equation}

\begin{equation}
\Delta E_{\mathrm{ads}} = \min(\Delta E_{\mathrm{i}})
\end{equation}

where \(E_{\mathrm{slab}+\mathrm{i}}\) is the total energy of the relaxed adsorbate--slab system, \(E_{\mathrm{slab}}\) is the energy of the clean slab, and \(E_{\mathrm{gas}}\) is the gas-phase adsorbate reference energy. Negative \(\Delta E_{\mathrm{ads}}\) values indicate stronger binding, while positive values indicate weaker adsorption. Because adsorption energies of key intermediates correlate with rate-determining steps, they are widely adopted as activity and selectivity descriptors.

Finding the adsorption energy \(\Delta E_{\mathrm{ads}}\), which is the global minimum energy across all configurations, demands performing numerous calculations over numerous surface terminations, spin states, and numerical settings.\cite{ock2026adsorb, lan2023adsorbml, chen2025multi} These costs have motivated machine learning (ML) surrogate models that approximate DFT at much lower computational expense while maintaining high accuracy.\cite{klawohn2023gaussian, kyvala2023optimizing, srinivasan2024electronic, li2022deep, thant2025kernel, pun2019physically, cao2024machine} Graph neural networks (GNNs), which encode atoms and bonds as graph nodes and edges, are especially effective for predicting atomic energies and forces\cite{deng2023chgnet, liao2023equiformerv2, batzner20223, batatia2022mace, choudhary2021atomistic, chen2019graph, wood2025family} and underpin machine-learned interatomic potentials (MLIPs) trained on large datasets of DFT calculations and ab initio molecular dynamics simulations.\cite{kabyldamolecular, maheshwari2025beyond, han2025distmlip} AdsorbML formalizes ML-accelerated adsorption evaluation through slab construction, adsorbate placement, GNN relaxation, and energy computation, achieving 87.36\% agreement with full DFT calculations and a 2000-fold speedup. Meta's FAIRchem initiative\cite{FAIRChem} extends this ecosystem with the UMA family of MLIPs\cite{wood2025family}, which provide transferable energy and force predictions across diverse chemistries.

AI agents offer a different set of capabilities for materials science. In this context, agency refers to the ability of an AI system to translate a high-level scientific objective into a sequence of actions, select and use external tools, observe the results, update its plan, and decide the next step. Transformer-based large language models (LLMs) provide the reasoning backbone for such systems,\cite{vaswani2017attention}, but agency arises from coupling this reasoning to tool use, memory, constraints and feedback. In scientific workflows, this enables models to move beyond one-shot prediction toward iterative hypothesis generation, experiment selection, and result interpretation.\cite{chaudhari2024alloybert, jadhav2025llm, badrinarayanan2025mofgpt, chandrasekhar2025nanogpt, chandrasekhar2024amgpt, ock2023catalyst, wang2022molecular} For catalyst screening and triage, an agentic workflow can retrieve structures, evaluate adsorption energetics, compare candidates against reaction-specific criteria, and refine subsequent choices based on quantitative feedback.

Recent studies show that agentic AI can accelerate scientific discovery by coupling language-based planning with literature, simulation and machine-learning tools.\cite{nigam2026polymeragentlargelanguagemodel, pak2026agentic, jadhav2026large, zeng2025llm, han2025tdflow, ock2025large, chandrasekhar2507automating, chaudhari2025modular, ghafarollahi2025rapid, wang2025swarms} In materials discovery, LLMatDesign\cite{jia2024llmatdesign} uses self-reflective LLM agents to modify and evaluate candidate materials with external tools, while LLEMA\cite{abhyankar2025accelerating} couples LLM generation with chemistry-informed evolutionary search for multi-objective design. In catalysis, CatGPT by Mok and Back\cite{mok2024generative} uses a fine-tuned GPT-2\cite{radford2019language} model to generate adsorbate-conditioned surface structures, with BERT-based\cite{devlin2019bert} validity filtering and EquiformerV2\cite{liao2023equiformerv2}/DFT validation. ChemReasoner by Sprueill et al.\cite{sprueill2024chemreasoner} casts screening as LLM-guided search with quantum-chemical feedback, beam search and context-aware planning.\cite{achiam2023gpt} Yao et al.\cite{yao2025llm} combine literature mining\cite{ye2024prompt}, supervised ML and SHAP interpretability\cite{lundberg2017unified} for MgH\(_2\) catalyst design, MASTER by Rothfarb et al.\cite{Rothfarb2025HierarchicalML} demonstrates a closed-loop agent for CO adsorption, and MAESTRO by Mok et al.\cite{mok2026reasoning} presents a multi-agent system for ORR single-atom catalyst discovery. These efforts highlight the growing role of agents in catalyst screening, but they remain limited by narrow reaction scope, direct structure generation, fragmented tools, dependency conflicts, or limited portability. We therefore use the Model Context Protocol (MCP)\cite{mcp} as a lightweight interface for tool-grounded agents, linking database retrieval, surface construction, ML relaxation, adsorption-energy evaluation, and iterative refinement within a single closed-loop workflow.

In this study, we introduce Catalyst-Agent, an autonomous AI agent that efficiently scales and accelerates catalyst screening by integrating tool-augmented LLM planning with ML-accelerated adsorption energy evaluation. Unlike prior approaches that focus on structure generation\cite{mok2024generative} or literature mining\cite{yao2025llm}, Catalyst-Agent retrieves candidate crystals from established databases, constructs physically consistent surface models, evaluates adsorption energetics using an MLIP, and iteratively refines promising candidates via structural modifications. All of this is orchestrated autonomously within a single closed-loop MCP-based framework. We evaluate Catalyst-Agent on three electrocatalytic screening tasks: the oxygen reduction reaction (ORR), nitrogen reduction reaction (NRR), and CO\(_2\) reduction reaction (CO$_2$RR), and demonstrate its effectiveness across chemically diverse search spaces.

\section*{Methods}
\vspace{0.5em}
\label{sec:Methods}
\begin{figure}[H]
    \centering

    \begin{subfigure}[t]{\textwidth}
        \centering
        \includegraphics[width=1.1\textwidth]{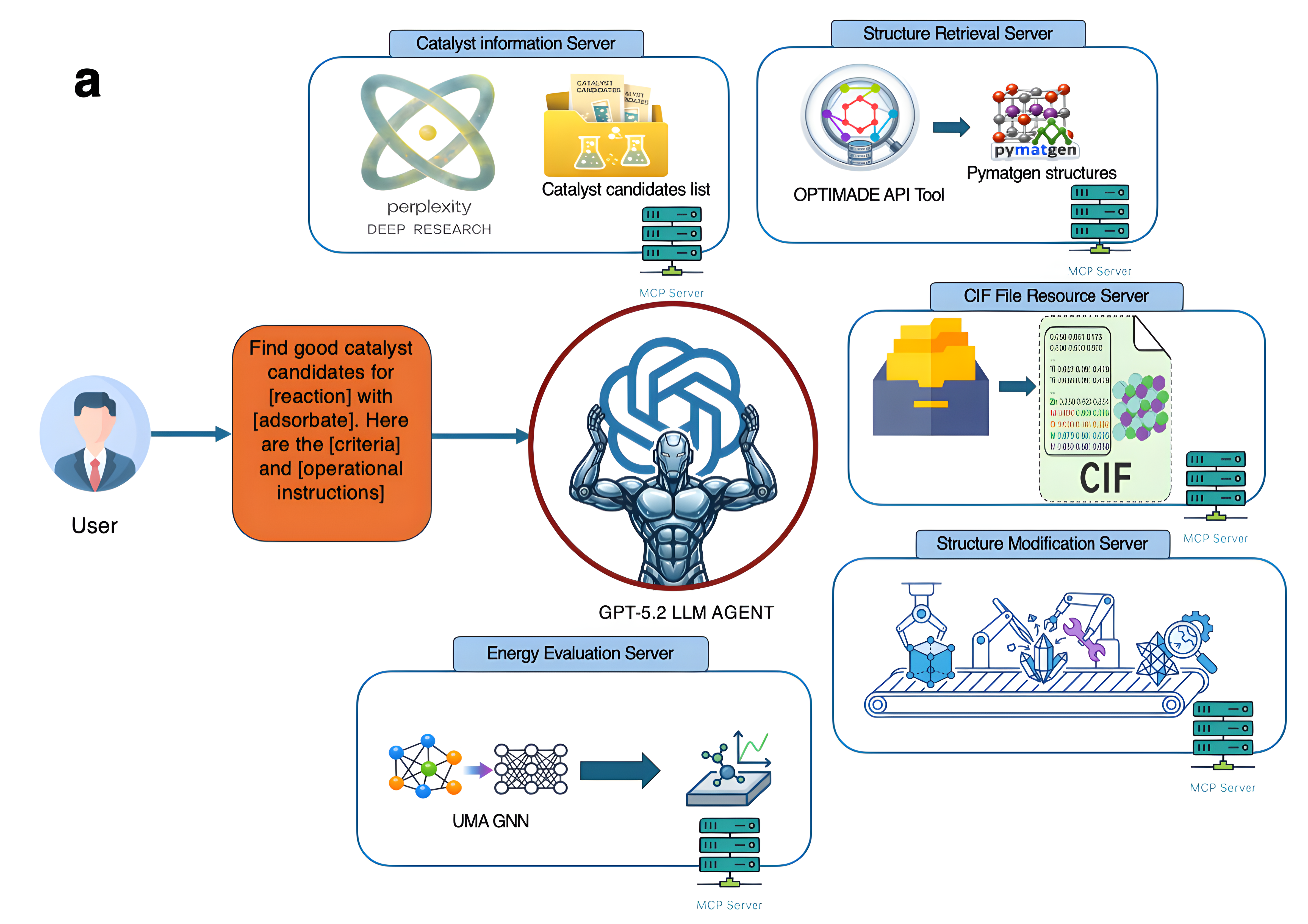}
       
        \label{fig:workflow:a}
    \end{subfigure}

    \begin{subfigure}[t]{1\textwidth}
        \centering
        \includegraphics[width=1\textwidth]{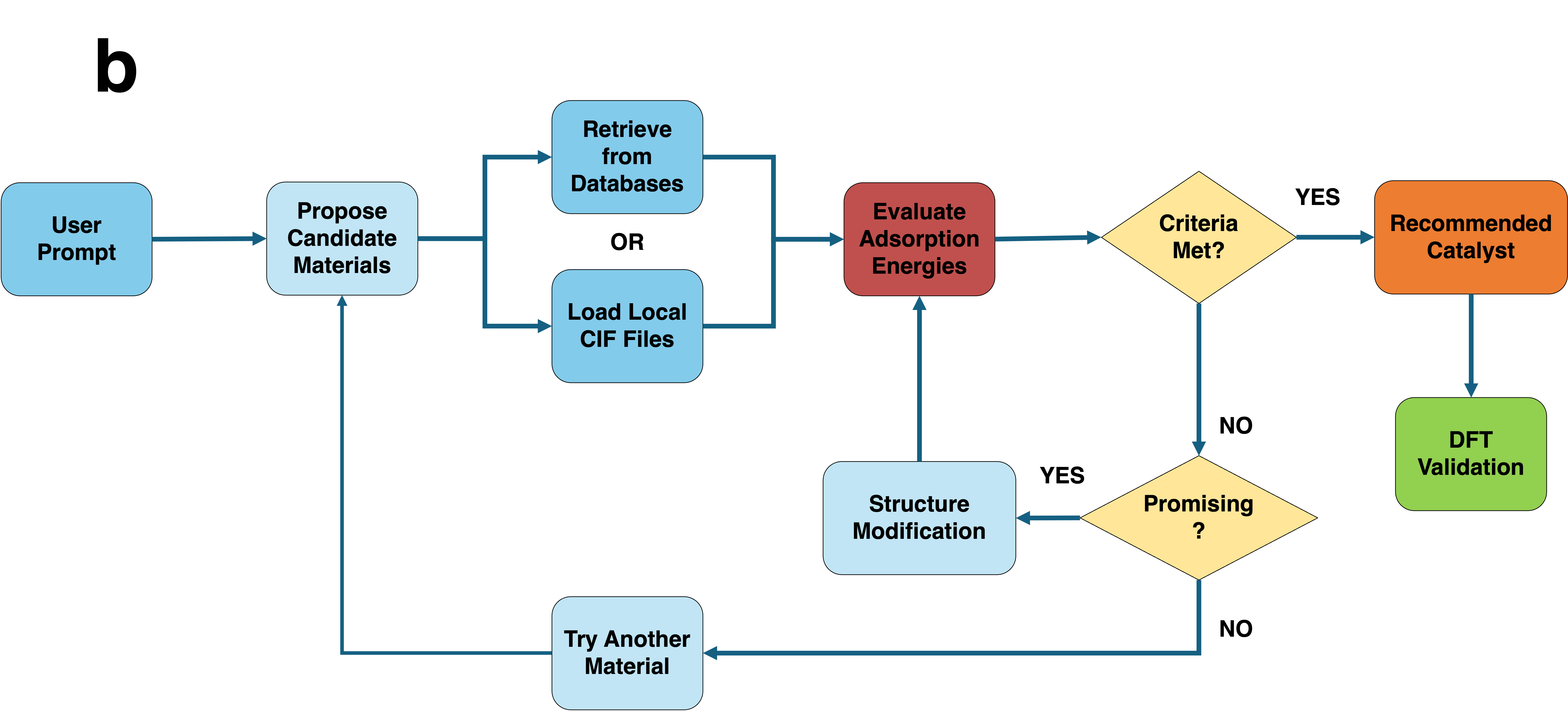}
        
        \label{fig:workflow:b}
    \end{subfigure}

    \caption{Overview of Catalyst-Agent: (a)~the LLM Agent supervisor coordinates actions while the five specialized MCP servers: Catalyst Information, Structure Retrieval, CIF File Resource, Structure Modification and Energy Evaluation, provide external functions (b)~the end-to-end user prompt-defined flowchart showing structure retrieval, modification, evaluation, and iteration.}
    \label{fig:workflow}
\end{figure}

\subsection*{Agentic framework and workflow}

Catalyst-Agent's main contribution is an operational framework, shown in Figure~\ref{fig:workflow}, where high-level screening goals serve as rewards to drive tool-grounded adsorption energy calculations. Rather than consolidating all capabilities into a single monolithic tool server, the framework distributes its functionality across five specialized MCP\cite{mcp} servers, each encapsulating a distinct stage of the catalyst screening pipeline. This is done to efficiently manage diverse and often conflicting dependencies and packages. An LLM-based agent, powered by GPT-5.2 running inside the Codex CLI environment\cite{codex_cli_2025}, acts as the central orchestrator: it receives a user prompt specifying a target reaction, adsorbates, and activity criteria, and then plans, sequences, and invokes the appropriate MCP server tools to execute the end-to-end screening workflow.

Each MCP server exposes a well-defined set of tools as shown in Table \ref{tab:mcp-servers-tools}, with predefined input schemas and structured JSON outputs, enabling the agent to invoke them through a request/response command-line interface and to feed the returned results directly into subsequent reasoning steps within its model context. A key architectural feature is subprocess isolation: the tool endpoints act as lightweight proxies that outsource computationally intensive tasks (database queries, GPU-accelerated relaxations) to external terminals, thereby preventing GPU overloading, enabling fine-grained error handling, and ensuring that a failure in one server does not crash the entire system.

The user-prompt driven end-to-end flowchart is depicted in Figure~\ref{fig:workflow}b. The user guidance prompt is passed to the agent, and it is empowered as the autonomous primary decision-maker to specify the operational requirements, target adsorbates, and descriptor criteria for success. Subsequently, the agent proceeds through the following stages: obtaining an initial candidate list, retrieving crystal structures, optionally applying material modifications, evaluating adsorption energies via the AdsorbML-UMA pipeline, checking whether the computed descriptors satisfy the acceptance criteria, and, if the result is a near-miss, iterating back to modify the structure and re-evaluate. The following subsections describe each of the five MCP servers in detail. An example of the MCP tool-call sequence and JSON outputs for a single material screening is provided in Section~S2, and a representative user guidance prompt is given in Section~S1.

\begin{table}[htbp]
\centering
\caption{MCP servers, exposed tools, and tool functions in the Catalyst-Agent framework.}
\label{tab:mcp-servers-tools}
\small
\renewcommand{\arraystretch}{1.25}
\begin{tabularx}{\textwidth}{@{}>{\raggedright\arraybackslash}p{3.4cm}>{\raggedright\arraybackslash}p{3.5cm}>{\raggedright\arraybackslash}X@{}}
\toprule
\textbf{Server} & \textbf{Tool} & \textbf{Function exposed to the agent} \\
\midrule
Catalyst Information (Perplexity Deep Research) & \texttt{deep\_research} & Searches web-based open-access literature for reaction-specific catalyst candidates and returns a seeded material list with supporting context. \\
Structure Retrieval & \texttt{optimade\_\allowbreak{}structure\_\allowbreak{}search} & Queries OPTIMADE providers such as the Materials Project and OQMD using composition constraints, then returns pymatgen structures and compact summaries for agent selection. \\
CIF File Resource & \texttt{list\_cif\_files} & Lists available CIF inputs for user-defined or Catalysis Hub structures, enabling direct structure selection when OPTIMADE retrieval is unsuitable. \\
Structure Modification & \texttt{create\_\allowbreak{}and\_\allowbreak{}serialize\_\allowbreak{}slab} & Creates serialized surface slabs and applies agent-selected modifications such as strain, vacancies or elemental substitutions before re-evaluation. \\
Energy Evaluation & \texttt{adsorbml\_evaluate} & Constructs requested facets, places adsorbates, and runs AdsorbML with UMA to report minimum adsorption energies and calculation diagnostics. \\
\bottomrule
\end{tabularx}
\end{table}

\subsubsection*{Catalyst Information Server}
\label{sec:results:mcp:catalyst-info}

The Catalyst Information Server provides the agent with an initial seeded list of candidate materials for a given reaction. For the ORR and NRR screening campaigns only, the candidate lists are generated using Perplexity's Deep Research\cite{perplexity_deep_research_2025}, which collects open-access, web-based literature-informed suggestions of promising materials. Perplexity is selected for its highly accurate literature review capability.\cite{rajamanicomparative} These lists are supplied to the supervisor LLM for further action. This server thus acts as the entry point of the screening pipeline, furnishing the agent with a curated starting pool of materials to explore. However, the agent is free to explore other materials not given to it by this server. Example material lists are referred to in Section~S3.

\subsubsection*{Structure Retrieval Server}
\label{sec:results:mcp:structure-retrieval}

The Structure Retrieval Server exposes tools that query the OPTIMADE API\cite{andersen2021optimade} to retrieve inorganic crystalline structures from multiple databases, for the ORR and NRR cases only. In this work, two providers are used: the Materials Project\cite{jain2013commentary} and the Open Quantum Materials Database (OQMD)\cite{saal2013materials, kirklin2015open}. The agent supplies element names and the number of distinct elements in the target formula, and the server returns matching structures. Two databases are chosen so that if one database's API endpoint fails, then the other will return structures. It is very rare for both endpoints to malfunction. Other OPTIMADE-accessible databases, such as the Alexandria Materials Database and the Crystallography Open Database\cite{gravzulis2012crystallography} are also accessible but were not queried in this study to limit computational overhead.

The server returns two complementary outputs: (i)~a JSON dictionary with complete, serializable pymatgen\cite{ong2013python} structure representations (formula, space group, lattice parameters, atomic positions and other structural metadata), and (ii)~a compact summary intended for perusal by the LLM supervisor, which reduces the amount of text, to simplify material selection and prevent overflow of the LLM's context window. The recorded structures can be converted to ASE Atoms objects\cite{larsen2017atomic} for compatibility with FAIRChem's AdsorbML in the adsorption energy evaluation stage\cite{lan2023adsorbml}.

\subsubsection*{CIF File Resource Server}
\label{sec:results:mcp:cif-server}

The CIF File Resource Server provides an alternative pathway for supplying crystal structures to the pipeline. Instead of querying online databases, the agent can obtain direct links to Crystallographic Information Files (CIF)\cite{hall1991crystallographic} provided locally by the user. When a CIF file is used, the server automatically performs symmetry analysis on the input structure with the symmetry precision threshold as $1.5\,\text{\AA}$, and logs the crystallographic space group symbol and number. This metadata is included in the returned results JSON alongside the parsed structure, for information tracking. The CIF pathway is particularly useful for materials that may not yet be indexed in OPTIMADE providers or for user-defined custom structures. Considering the rarity of stable surfaces and viable materials for the CO\(_2\) reduction reaction, the initial CIF files for CO\(_2\)RR are instead obtained from the Stanford SUNCAT surface reaction database, known as Catalysis Hub\cite{winther2019catalysis} and supplied through this server. Manual filtering is employed, as many reaction-specific material entries are missing in this database.

\subsubsection*{Structure Modification Server}
\label{sec:results:mcp:structure-modification}

The OPTIMADE search and AdsorbML workflow described above permit the agent to evaluate adsorption energetics on slabs generated from structures available in external materials databases. While this represents a useful starting point, it restricts the evaluation of possible catalyst surfaces to the finite set of parent structures found within the external databases and does not permit the agent to explore structural modifications to otherwise suitable candidate structures. To address this limitation, we developed a Structure Modification Server. This provides a reproducible method to extend the initial search space beyond the fixed set of structures contained in the external databases and to assess whether commonly employed catalyst optimization techniques, including surface substitution, vacancy creation, and elastic deformation, could enhance the adsorption energetics of structures that represent reasonable candidates based upon a preliminary assessment. By restricting the agent's ability to make the above-defined modifications, we ensure that the search process remains transparent: each modification has well-defined inputs, is applied through a deterministic geometric transformation, and is stored in the resulting output JSON file for future reference. Prior to performing the adsorption-energy calculation, and for each requested Miller-index, the parent structure is transformed into a slab by applying the same slab-generation procedure used for the unmodified calculation. The modification server then applies transformations directly to the previously created slab. Surface layers are determined by sorting all slab atoms along the surface normal, taken as the Cartesian $z$ direction, according to their $z$-coordinate values. Layers are then assigned as groups of atoms whose $z$-coordinate values differ by no more than 0.35~\AA{} from the mean height of the current layer. Group 1 consists of the surface layer, and Group 2 consists of the subsurface layer. This layer-based description gives the agent a consistent means to identify surface versus subsurface atoms in a manner independent of manual assignment of site labels for every slab.

Substitution is used to model local variations in surface composition, as in alloying or doping studies where the electronic environment of adsorption sites is manipulated.\cite{sun2024machine} A substitution is made only if the agent identifies both a target element ($X_{\text{target}}$), representing the original chemistry at an adsorption site, and a replacement element ($X_{\text{dopant}}$). Only atoms in the top surface layer that correspond to $X_{\text{target}}$ are considered for substitution. For single-site substitution, the atom with the highest $z$ value among the candidate atoms is chosen for substitution, and its symbol is changed to that of $X_{\text{dopant}}$, while its atomic position is maintained. For two-site substitution, two distinct surface atoms are again chosen from the set of candidate atoms based upon a deterministic ``farthest-site`` rule. The first selected atom corresponds to the highest-$z$ target atom. The second selected atom corresponds to the target atom that maximizes the Cartesian distance to the first selected atom. If fewer than the required number of target atoms are present in the surface layer, then no substitution occurs, and this fact is noted in the results.

Vacancies constitute an alternative means to examine how sensitive adsorption energies are to undercoordinated surface environments and missing-atom defects, which are characteristic features of many catalyst surfaces.\cite{hu2021sulfur} Vacancy operations can be performed on both surface and subsurface layers. For a surface vacancy operation, an atom is selected from the top layer and for a subsurface vacancy operation, an atom is selected from the second layer below the surface. An optional target element can be provided by the agent to limit which atoms are eligible for removal during vacancy operations. The server then selects for removal the highest-$z$ atom in the appropriate layer that satisfies any specified elemental filtering criterion. If no qualifying atom exists within the selected layer, no atoms are removed and an appropriate message is written to the output indicating why no removal occurred. When vacancy and substitution operations are both requested, vacancies are applied prior to substitution, since subsequent dopant selections are performed on the structurally modified slab that is passed to AdsorbML. 

Strain can also be applied uniformly by the agent throughout both in-plane directions of the slab. Strain is widely recognized as a descriptor-level perturbation used to influence adsorption energies.\cite{oliveira2025strain} The values of both the a and b lattice dimensions are scaled down uniformly by the factor (1-$\epsilon$), where $\epsilon$ $> 0$ represents compressive strain and $\epsilon$ $< 0$ represents tensile strain. The value of the c lattice dimension remains unaltered. 

After any requested modification is applied, the modified slab is passed to the same AdsorbML workflow used for the parent surface. AdsorbML generates adsorbate placements, relaxes the adsorbate-slab structures, and returns the minimum referenced adsorption energy together with the number of valid configurations and detected anomalies. The modification metadata, including substituted atom indices, removed atom indices, target and replacement elements, strain magnitude, and any failed modification attempts, is stored alongside the adsorption results. This bookkeeping provides context for the agent's subsequent decisions and allows the final screening trajectory to be traced back to the exact structural modifications evaluated.

\begin{figure}[H]
\centering
\begin{subfigure}[t]{0.46\textwidth}
  \centering
  \includegraphics[width=1.1\linewidth]{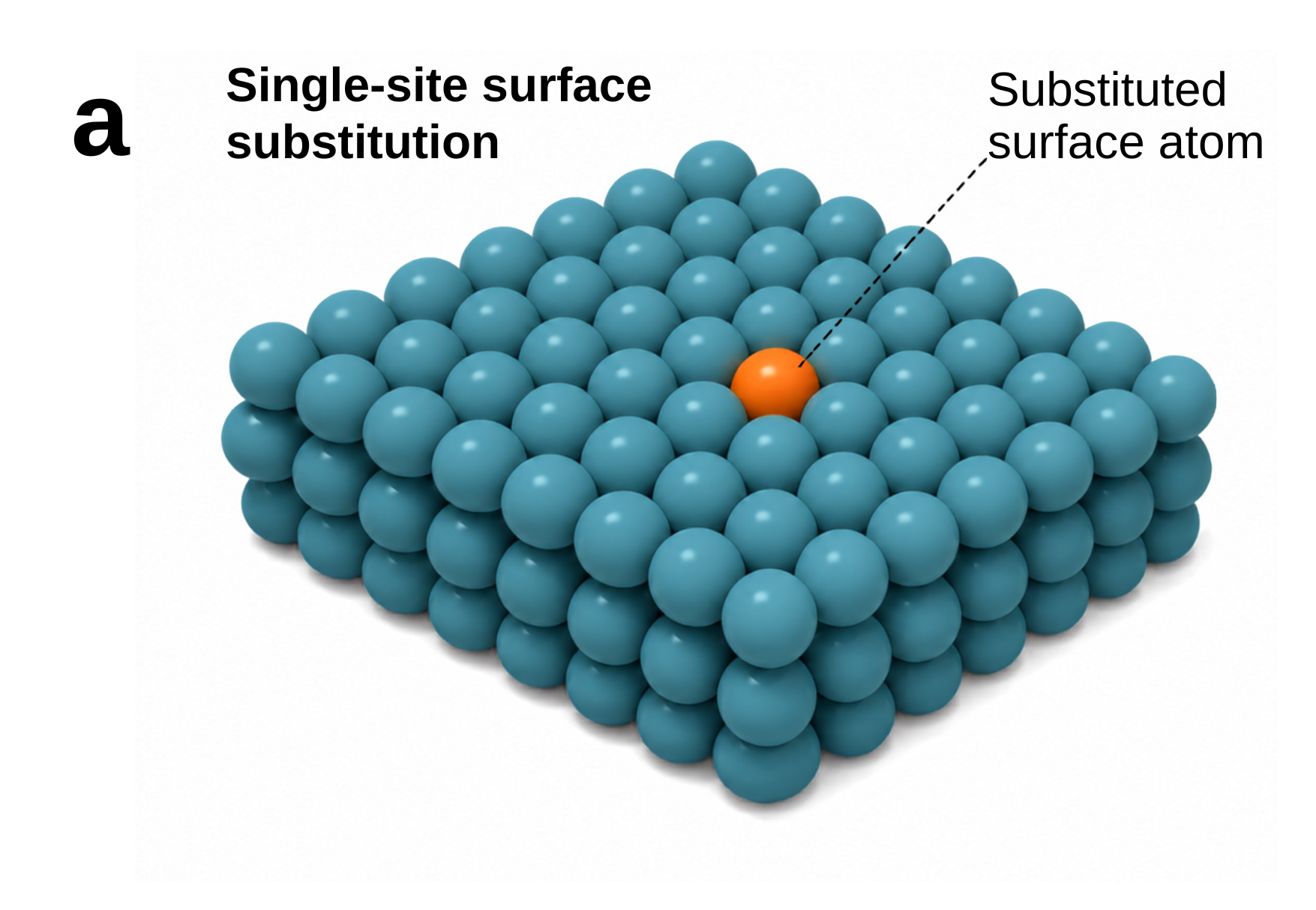}
  \label{fig:surface_mods_single_site}
\end{subfigure}
\hfill
\begin{subfigure}[t]{0.44\textwidth}
  \centering
  \includegraphics[width=1.1\linewidth]{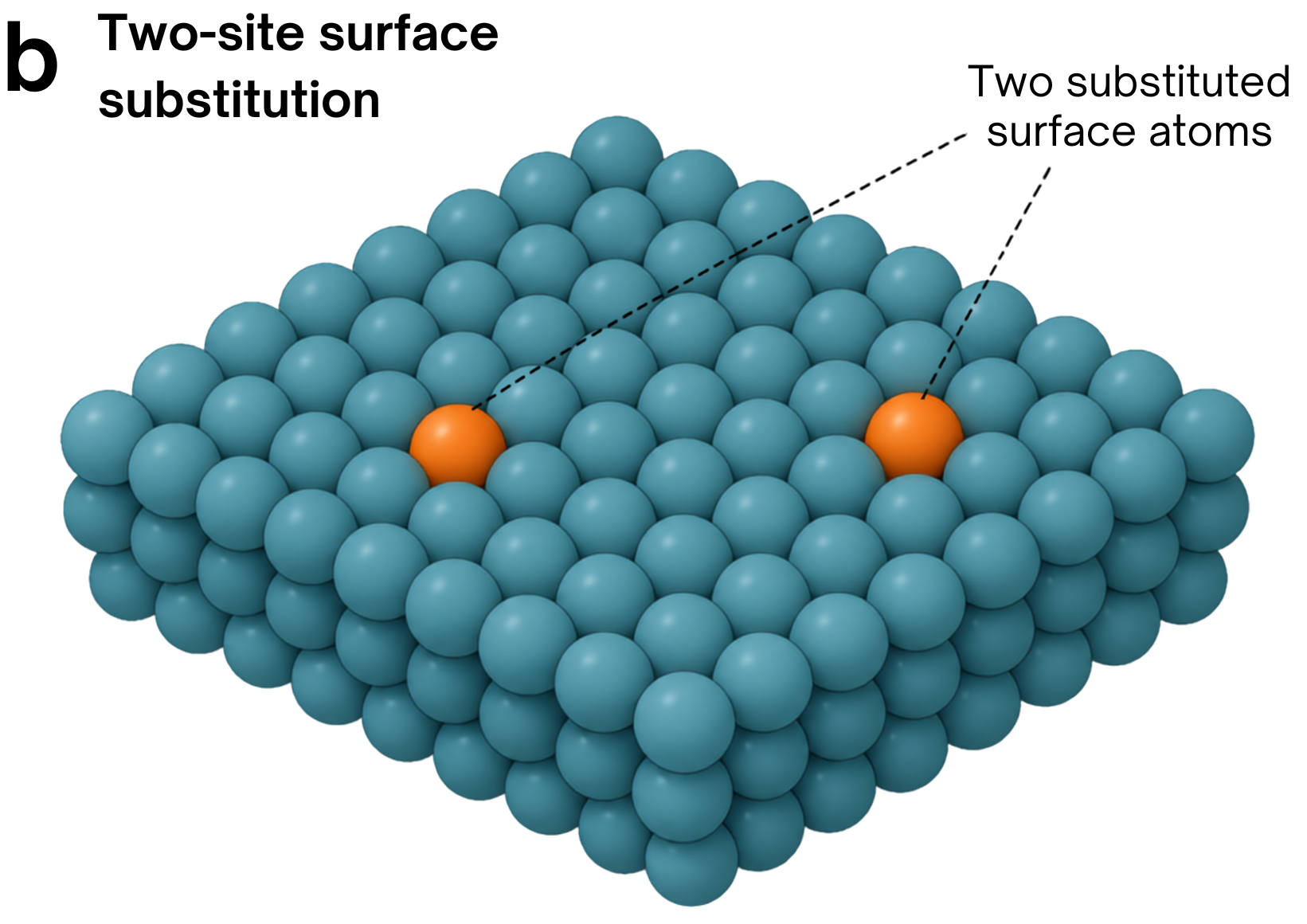}
  \label{fig:surface_mods_two_site}
\end{subfigure}

\vspace{0.5em}

\begin{subfigure}[t]{0.44\textwidth}
  \centering
  \includegraphics[width=1.1\linewidth]{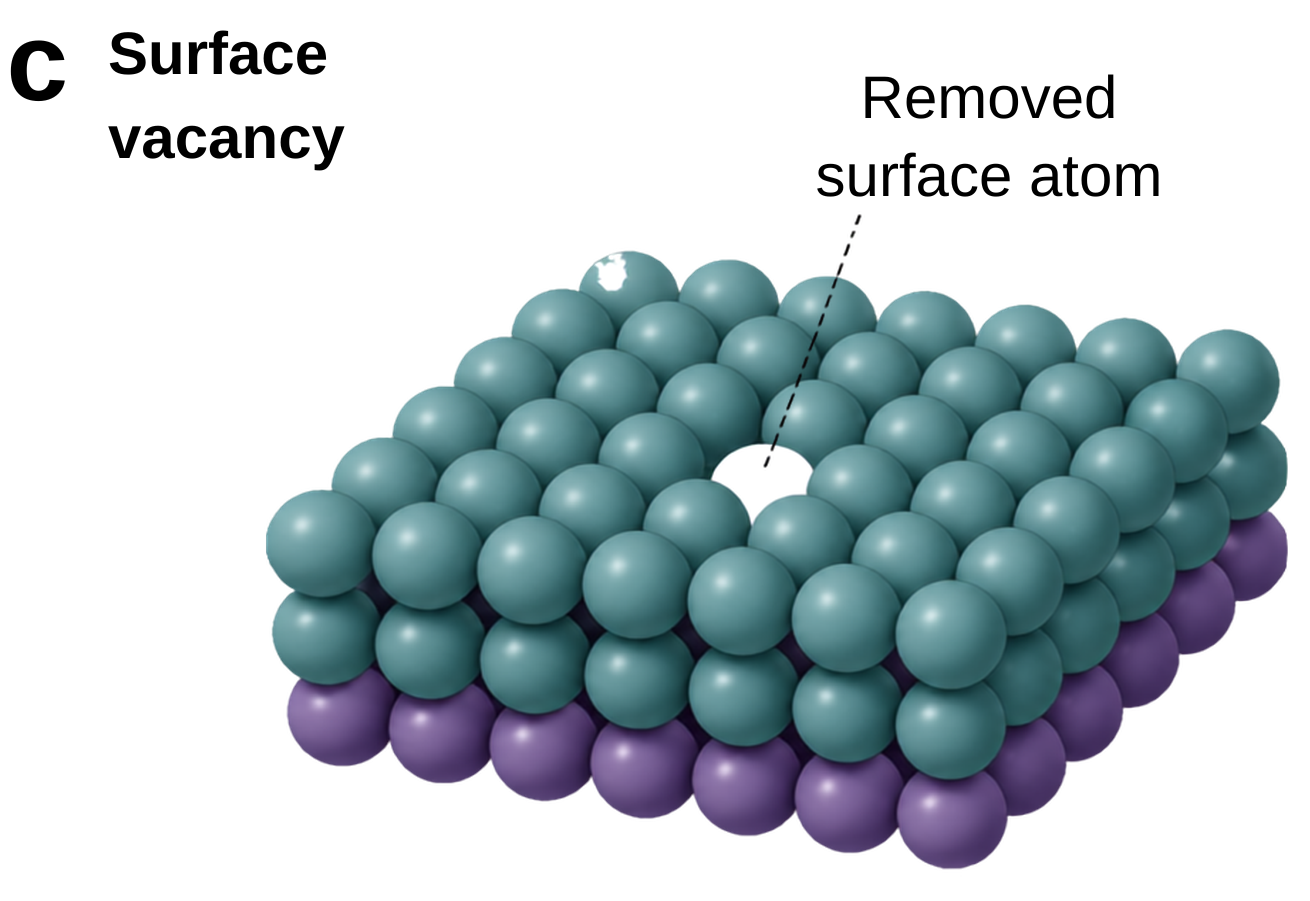}
  \label{fig:surface_mods_surface_vacancy}
\end{subfigure}
\hfill
\begin{subfigure}[t]{0.44\textwidth}
  \centering
  \includegraphics[width=1.1\linewidth]{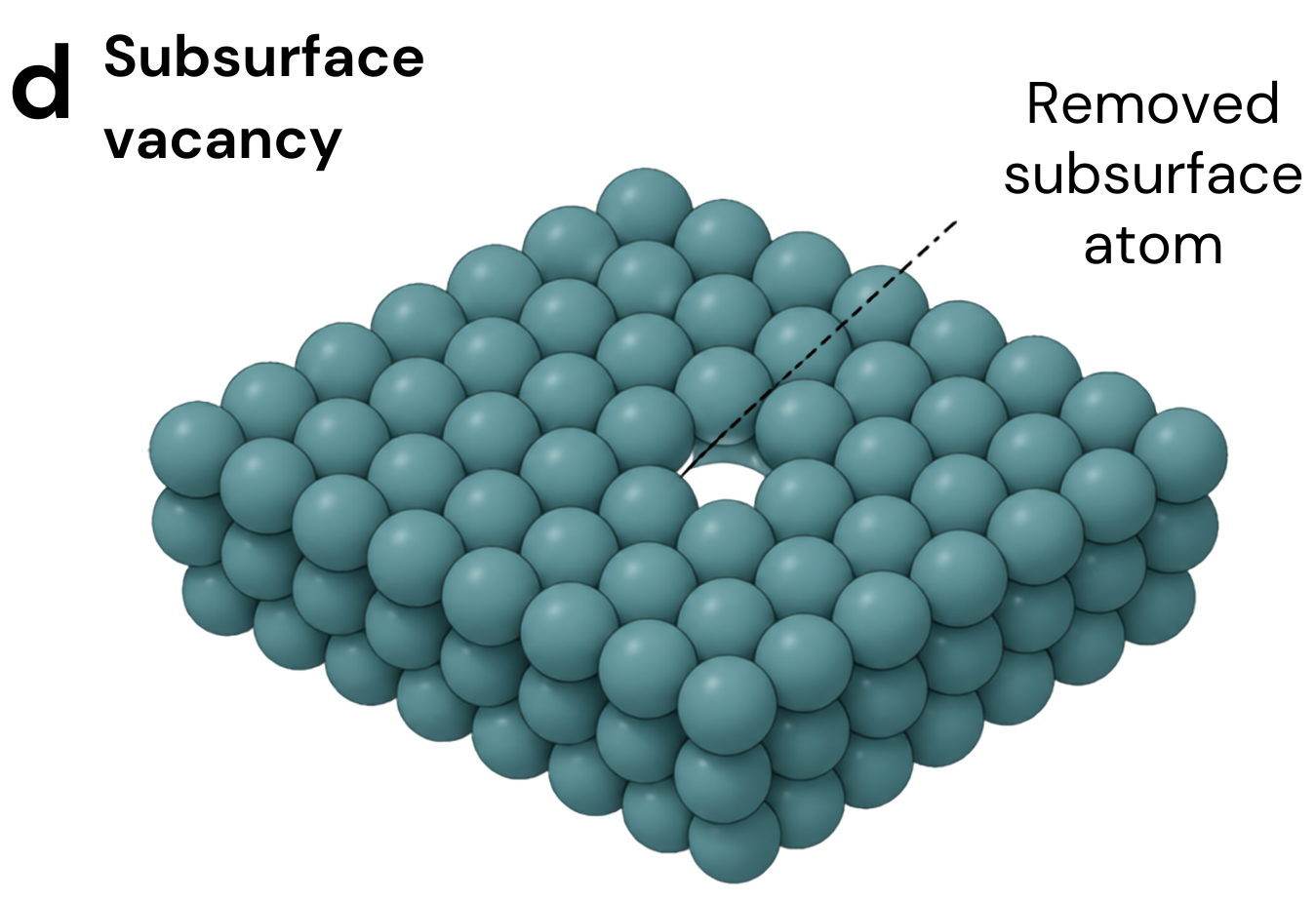}
  \label{fig:surface_mods_subsurface_vacancy}
\end{subfigure}

\vspace{0.5em}

\begin{subfigure}[t]{0.64\textwidth}
  \centering
  \includegraphics[width=\linewidth]{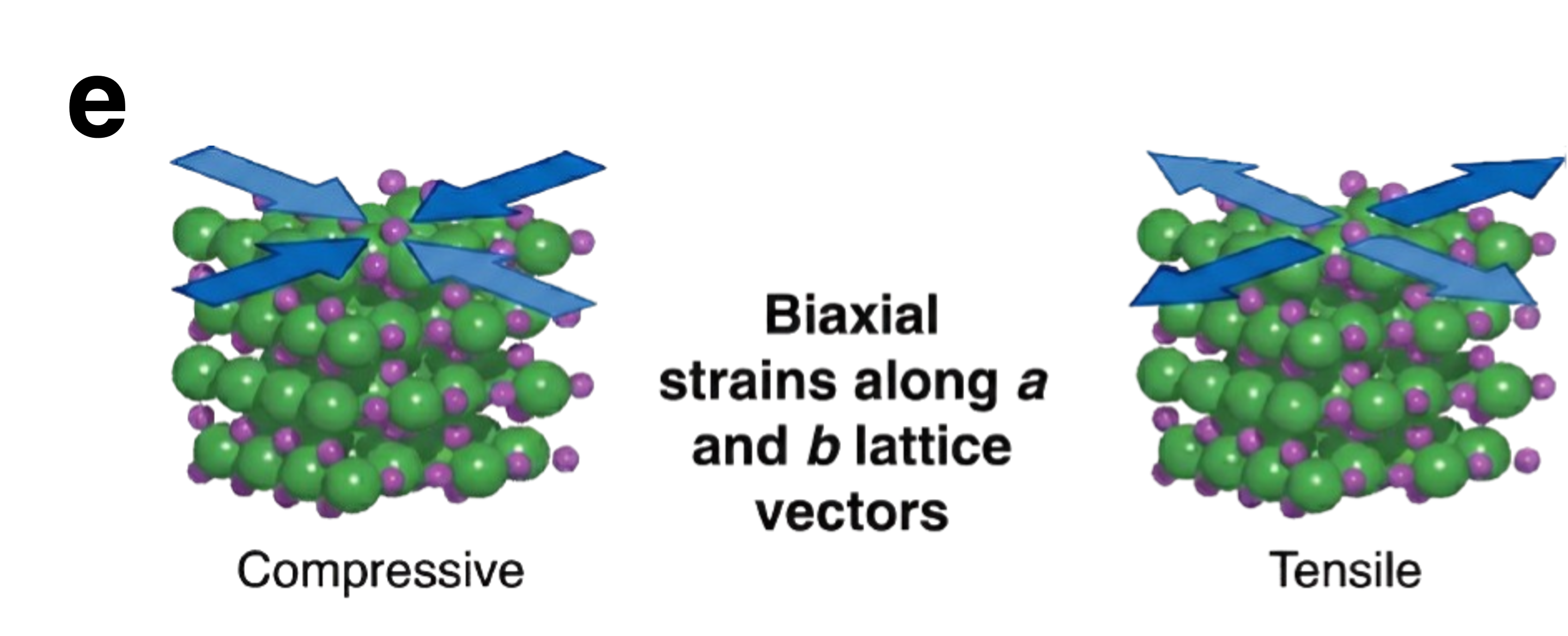}
  \label{fig:surface_mods_strain}
\end{subfigure}

\caption{Available structural modifications after slab construction:
(a) one-site surface substitution,
(b) two-site surface substitution,
(c) surface vacancy,
(d) subsurface vacancy and (e) in-plane strain}
\label{fig:surface_mods}
\end{figure}

\newpage
\subsubsection*{Energy Evaluation Server}
\label{sec:results:mcp:energy-eval}

The Energy Evaluation Server is the main computational component of Catalyst-Agent. It takes a candidate structure (either from the Structure Retrieval Server's output, a CIF file from the CIF File Resource Server, or an ASE Atoms object from the modification server) along with an adsorbate label and up to five Miller indices per tool call, and will evaluate the adsorption energy on all requested facets in parallel. In this way, Catalyst-Agent's throughput is greatly enhanced to provide a ``Multi-Facet Snapshot`` of the surface chemistry of a material.

Internally, for each requested Miller-index, the server constructs an oriented slab that satisfies minimum lateral size requirements. The minimum in-plane tiling threshold (\texttt{min\_ab}) is chosen adaptively: $8.0$~\AA{} for low-index facets ($\max(|h|,|k|,|l|)\le 3$) and $6.0$~\AA{} otherwise, improving robustness for high-index surfaces. Detailed debug information is returned if slab tiling fails, as an update to the LLM supervisor agent.

Adsorption energies are calculated via the AdsorbML workflow using the UMA small (uma-s-1p1) MLIP to relax multiple initial adsorbate placements on each slab. The relaxations are performed using a quasi-Newton optimizer (LBFGS) with predetermined convergence criteria (force tolerance  \texttt{fmax=0.05~eV/\AA} and \texttt{number of steps=300}). Each evaluation returns a structured JSON summary containing: (i)~the total number of attempted configurations, (ii)~the number of valid configurations, (iii)~the number of detected anomalies, (iv) structure reference details and (v)~the global minimum energy prediction among the valid runs, including energy splits that decompose the total into slab, gas-phase reference and adsorbate-slab contributions. A total of five adsorbate placements are used for an initial screening across all materials to determine successful candidates for the first round. This is a pre-defined setting not subject to change by the agent. Candidates from the first round are then subjected to a second round of thirty placements to perform a refined evaluation of the successful candidates and to finalize the recommendations for the best-performing catalysts.

To make the workflow fault-tolerant, the MCP tools act as lightweight proxies rather than running every calculation inside the same process as the agent. For each requested facet, the Energy Evaluation server launches a new Python subprocess to perform the slab construction, adsorbate placement, and UMA relaxation for that facet. If the facet fails due to an issue with the slab, problems during relaxation, or other runtime errors, it would be terminated but the remaining facet computations could proceed independently. Thus, if there was a problem with evaluating one particular facet at the time when multiple facets are being evaluated, it would not prevent the evaluation of those additional facets from proceeding.

\subsubsection*{Workflow summary}

The high-level overview of the process orchestrated by the agent across these five servers is: propose candidate chemistries (Catalyst Information Server) $\rightarrow$ retrieve crystal structures (Structure Retrieval Server or CIF File Resource Server) $\rightarrow$ evaluate adsorption energies on multiple facets (Energy Evaluation Server) $\rightarrow$ compare to descriptor criteria $\rightarrow$ if the candidate is a near-miss but promising, apply strain or substitution doping (Structure Modification Server) $\rightarrow$ re-evaluate (Energy Evaluation Server), iterating until the activity criteria are met or the candidate base material is deemed unsuitable for further modification as shown in the subsequent section on agentic adaptive search. The agent is prompted to document its reasoning behind material selection and any modifications applied.

\subsection*{Adsorption tasks and descriptors}
\label{sec:results:targets-descriptors}

Catalyst-Agent was evaluated on three representative screening tasks. Each of these reactions is defined by a set of adsorbates and quantitative activity descriptors. All these quantities are obtained from the energy evaluation server, summarized by the minimum-energy valid configuration for a given material-adsorbate system. Table~\ref{tab:targets-descriptors} summarizes the targeted reactions, the adsorbates evaluated, and the descriptors used for decision making.

\begin{table}[H]
\centering
\caption{Targeted reactions, AdsorbML adsorbates, and descriptor criteria used by Catalyst-Agent 
in each screening campaign. Energies are in eV.}
\label{tab:targets-descriptors}
\small
\renewcommand{\arraystretch}{1.4}
\begin{tabularx}{\textwidth}{@{}
    >{\raggedright\arraybackslash}p{3.5cm}
    >{\centering\arraybackslash}p{2.8cm}
    >{\raggedright\arraybackslash}X
    >{\raggedright\arraybackslash}X
@{}}
\toprule
\textbf{Task} & \textbf{Adsorbates} & \textbf{Descriptor(s)} & \textbf{Target Criterion} \\
\midrule
Oxygen Reduction Reaction (ORR) &
    $*$OH &
    $E_{\mathrm{ads}}(*\mathrm{OH})$ &  $E_{\mathrm{ads}}(*\mathrm{OH})\in[0.9,\,1.1]$  
    \\[4pt]

\ce{CO2} Reduction to CO (CO$_2$RR) &
    $*$CO,\;$*$H &
    $E_{\mathrm{ads}}(*\mathrm{CO}),\;E_{\mathrm{ads}}(*\mathrm{H})$ &
    $E_{\mathrm{ads}}(*\mathrm{CO})\in[-0.82,\,-0.52]$ and
    $E_{\mathrm{ads}}(*\mathrm{H})\geq 0.33$ \\[4pt]

Nitrogen Reduction (NRR) &
    $*$N,\;$*$N$_2$ &
    $E_{\mathrm{ads}}(*\mathrm{N}),\;E_{\mathrm{ads}}(*\mathrm{N}_2),\;\Delta$ &
    $E_{\mathrm{ads}}(*\mathrm{N})\in[-2.0,\,-0.5]$ and
    $\Delta = E_{\mathrm{ads}}(*\mathrm{N}) - E_{\mathrm{ads}}(*\mathrm{N}_2) > 0$ \\
\bottomrule
\end{tabularx}
\end{table}

The choices of descriptors here are obtained from the literature. For oxygen electrocatalysis, *OH adsorption energy readily provides an approximate indicator of binding strength, and the optimal band of (0.9-1.1 eV) is chosen for proximity to the adsorption energy of *OH on the Pt (111) surface.\cite{kulkarni2018understanding} Scaling relationships are given in great detail by Man et al.\cite{man2011universality} and similarity of mechanism to Pt is noted by Wang et al.\cite{wang2013oxygen}. For CO$_2$RR, the multiple conditions for the adsorption energies of *CO and *H give a dual constraint: *CO must bind strongly enough to form but not so strongly as to poison the catalytic surface, while \(E_{ads}\)*H must be sufficiently unfavorable to suppress the competitive hydrogen evolution reaction (HER) under the same conditions. These descriptors are the same as those utilized by Song et al.\cite{song2025inverse}. For the NRR case, the atomic-molecular descriptor $\Delta$ identifies surfaces that stabilize the atomic nitrogen intermediate (*N) relative to adsorbed \(N_{2}\), and simultaneously, a moderate \(E_{ads}\)*N window is applied to avoid non-binding and irreversible trapping.\cite{araujo2023high} Naturally, Catalyst-Agent is able to perform iterative strain and doping substitution tests for promising candidates that fall near, but not quite inside the acceptable boundaries. The complete raw screening outputs for all three reactions are referred in Section~S3.

\subsection*{AdsorbML adsorption-energy workflow}
\label{sec:methods:uma}

Adsorption energies were computed using FAIRchem’s AdsorbML workflow.\cite{wood2025family} AdsorbML generates candidate adsorbate--surface configurations by enumerating initial adsorbate placements on the catalyst slab surface using heuristic and random surface-site sampling strategies. In the heuristic component, candidate adsorption sites are generated from the slab surface geometry, while in the random component, additional surface sites are sampled to broaden the configurational search space. For each generated site, the adsorbate is placed on the surface with varied orientations, including rotations about the surface-normal direction and small tilts about the in-plane axes. Unphysical initial structures, such as adsorbates placed away from the slab or too deeply into the surface, are discarded prior to relaxation.

In this work, a two-stage AdsorbML sampling protocol was used to balance computational cost and configurational coverage. First, an initial screening calculation was performed using five AdsorbML-generated initial placements for each material-adsorbate-facet system. Systems that exhibited favorable adsorption energies in this initial screen were then shortlisted and re-evaluated using a denser set of 30 AdsorbML-generated initial placements.

All retained initial configurations were relaxed using the UMA small model. After relaxation, configurations were screened for AdsorbML anomalies, including adsorbate desorption, adsorbate dissociation, and significant adsorbate-induced surface reconstruction. Only valid relaxed configurations were retained for adsorption-energy evaluation.

For each facet, the workflow returns the number of attempted initial configurations, the number of valid relaxed configurations, the number of anomalous configurations, and the minimum-energy adsorbate-slab structure, along with its energy decomposition into clean-slab, gas-phase reference, and adsorbate-slab total energies. Typical wall times were approximately 8--12 min per facet on an NVIDIA A6000 GPU with 48 GB of memory. More information is given in Section~S4.

\subsection*{Model Context Protocol (MCP) tool servers}
\label{sec:methods:mcp}

The Model Context Protocol, or MCP, introduced by Anthropic in 2024 and often described as the “USB-C of AI,” provides a standard interface for connecting LLMs with tools, data sources, and workflows.\cite{mcp} Its architecture includes a host, client, and server: the host is the user-facing LLM application, the client manages communication, and servers provide tools, resources, and prompt templates. MCP has been adopted across major AI platforms and developer tools, contributing to its rapid growth and use in recent LLM-based systems.

\subsection*{LLM agent environment: Codex CLI}
\label{sec:methods:codex}

Codex CLI was used as the LLM command-line environment for planning and orchestration.\cite{codex_cli_2025} It was developed by OpenAI, primarily for agentic coding and can be run locally from the terminal. It is open-source software built using the Rust programming language.\cite{bugden2022rust} Codex CLI was used to run the GPT-5.2 model from OpenAI. This model was chosen for its excellent scientific understanding and reasoning capabilities. It was used with default settings provided by the API.

\begin{figure}[H]

\includegraphics[width=1.0\linewidth]{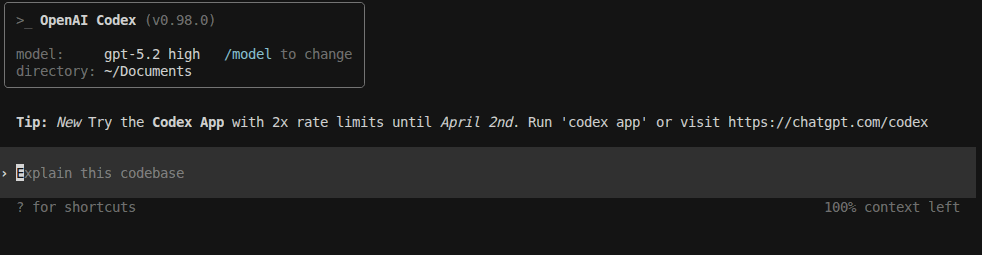}
\caption{Codex CLI (command-line interface) was used as the GPT-5.2 powered LLM-agent environment and user interface.
}
\label{fig:react}
\end{figure}

\subsection*{DFT Calculations and Validation Assessment}
\label{sec:si:dft-relaxations}

In order to validate the adsorption energy predictions made by Catalyst-Agent for certain materials, we completed several single-point DFT calculations. This validation involved comparing adsorption energies predicted by Catalyst-Agent with those produced by high-fidelity electronic-structure calculations. As such, this validation did not attempt to re-screen the entire candidate pool, but rather tested whether the ML-relaxed lowest-energy structures produced by Catalyst-Agent would be classified similarly when assessed using DFT.

To perform the validation for each material-adsorbate pair considered in this study, the same surface facet that was used in the Catalyst-Agent screening was created using FairChem's slab-generation tools. The initial placement of the adsorbate on the surface was created and optimized with AdsorbML using the small model. From that process, one AdsorbML-created placement was chosen as the starting point for further single-point DFT calculations. The adsorbate-slab combination, along with its clean counterpart and a gas-phase version of the adsorbate, was then optimized consistently with Quantum ESPRESSO\cite{giannozzi2009quantum}. The adsorption energy was calculated as follows:
\begin{equation}
E_{\mathrm{ads}} = E_{\mathrm{slab+adsorbate}} - E_{\mathrm{slab}} - E_{\mathrm{gas}}
\end{equation}

All calculations employed periodic boundary conditions and conditions were as close as possible to the OC20 dataset. The PSLibrary\cite{dal2014pseudopotentials} provided PAW-style UPF pseudopotentials, and the RPBE exchange--correlation functional was used. A wavefunction cutoff value of 350~eV (25.724~Ry) and a density cutoff value equal to eight times the wavefunction cutoff value were specified. To optimize the k-point sampling of the Brillouin zone for slabs, a Monkhorst--Pack mesh based on the in-plane dimensions of the cell was constructed in a similar manner to the OC20 dataset. An electronic smearing method based on Gaussians was applied with a smearing parameter of 0.1~eV for numerical stability. The electronic self-consistency convergence criterion was established at $10^{-4}$~eV.

No spin polarization or dispersion corrections were incorporated into either the electronic-structure or potential-energy calculations, to keep them similar to the OC20 dataset calculations. Ionic relaxation using the ASE interface to Quantum ESPRESSO utilized a conjugate-gradient optimization algorithm. Ionic relaxation convergence was defined by a maximum force below 0.03~eV~\AA$^{-1}$. Ionic relaxation iterations were limited to 200 steps.

\section*{Results and Discussion}

\subsection*{Illustrative example of agentic adaptive search}
\label{sec:results:agentic-reasoning}

\subsubsection*{\ce{CoGa2O4} (mp-765466) for ORR}
\label{sec:si:coga2o4-orr-adaptive-search}

\begin{figure}[H]
\centering
\includegraphics[width=1.0\linewidth]{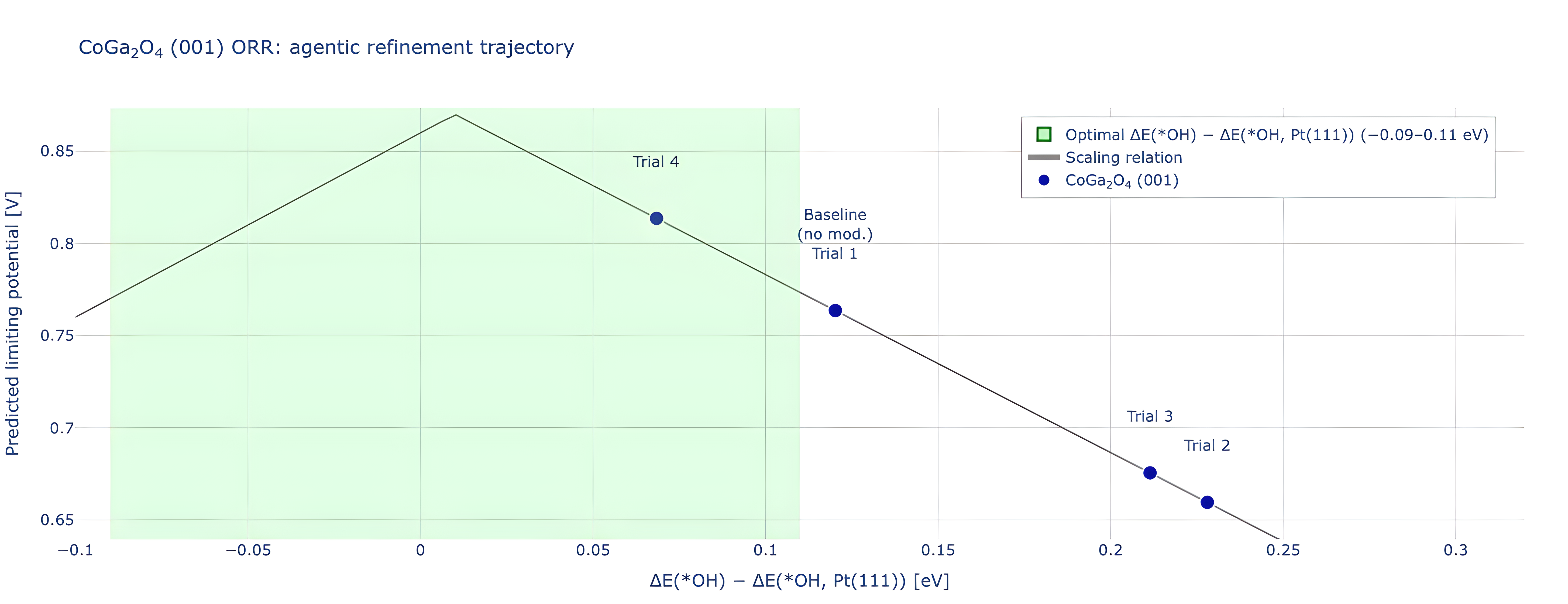}
\caption{Volcano plot of predicted limiting potential versus \(\Delta E(\ast\mathrm{OH}) - \Delta E(\ast\mathrm{OH}, \mathrm{Pt}(111))\) for \ce{CoGa2O4} (001) along the agent's refinement trajectory. \(\Delta E(\ast\mathrm{OH}, \mathrm{Pt}(111))\) is 0.99 eV. The green band marks the optimal window.}
\label{fig:cogao4-volcano}
\end{figure}

To illustrate how Catalyst-Agent autonomously goes through the closed-loop screening process, we show the complete history for the cobalt gallate spinel \ce{CoGa2O4} (Ga\(_2\)CoO\(_4\), Materials Project ID mp-765466) in the ORR screening campaign. This example is representative of the agent's iterative hypothesis--test--revise workflow and highlights how it combines chemical intuition with quantitative feedback from the energy evaluation server to converge on a successful modification strategy (Figure~\ref{fig:cogao4-volcano}).

\paragraph{Material selection rationale.}
\ce{CoGa2O4} is a normal spinel in which Co\(^{2+}\) occupies tetrahedral A-sites and Ga\(^{3+}\) fills octahedral B-sites. The agent selected it as a candidate because of the proven ORR activity of cobalt-based spinels (e.g.\ \ce{Co3O4}, \ce{MgCo2O4}, \ce{CoAl2O4}) and the opportunity to tune Co--O covalency by substituting the B-site cation (Al \(\leftrightarrow\) Ga). The initial baseline evaluation on the unmodified structure returned \(E_{\mathrm{ads}}(\ast\mathrm{OH}) = 1.1102\)~eV on the (001) facet (Trial 1), placing it barely above the optimal 0.9--1.1~eV target window. This made \ce{CoGa2O4} a promising candidate for iterative refinement.

\begin{figure}[H]
\centering
\includegraphics[width=0.4\linewidth]{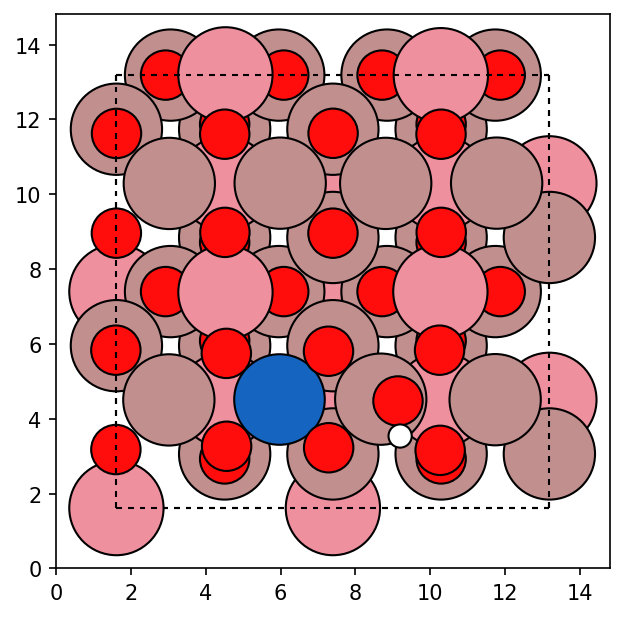}
\caption{Surface structure illustrating Ga\(\to\)Al substitution (aluminum atom in blue) with 2\% compressive strain, as evaluated in Trial~2.}
\label{fig:substitution}
\end{figure}

\paragraph{Trial 2 -- Ga\(\to\)Al surface doping with 2\% compressive strain:}
The agent's first hypothesis was that replacing the surface Ga atom with Al would increase local ionicity and simultaneously applying 2\% compressive strain would contract the Co--O bond lengths, together softening the slightly-too-strong \(\ast\)OH binding, as shown in Figure \ref{fig:substitution}. AdsorbML returned \(E_{\mathrm{ads}}(\ast\mathrm{OH})\) values of 1.2180~eV on (001) and 1.2042~eV on (100), both further from the optimal window than the baseline. The remaining facets were either strongly overbinding ((110): \(-0.6318\)~eV, (111): \(-0.8609\)~eV) or far too weak ((210): 0.0084~eV). The agent concluded that Al substitution combined with compression pushed adsorption in the wrong direction and abandoned this modification.

\paragraph{Trial 3 -- 2\% compressive strain only:}
To disentangle the dopant effect from the strain effect, the agent removed the Al substituent and re-evaluated the surface under 2\% compressive strain alone. The (001) facet returned 1.2014~eV and (100) returned 1.1159~eV, both still above the target window, while (111) showed extreme overbinding at \(-4.4015\)~eV. The agent inferred that compressive strain, regardless of doping, strengthens rather than softens \(\ast\)OH adsorption on the low-index spinel terminations of \ce{CoGa2O4}. This observation prompted the agent to reverse its strain strategy.

\paragraph{Trial 4 -- 2\% tensile strain only:}
Having established that compression moves adsorption away from the target, the agent applied 2\% tensile strain without any dopant. The (001) facet returned \(E_{\mathrm{ads}}(\ast\mathrm{OH}) = 1.0584\)~eV, placing it squarely inside the optimal 0.9--1.1~eV band. The agent successfully identified \ce{CoGa2O4} (001) under 2\% tensile strain as a viable ORR catalyst candidate after a total of four evaluations (one baseline and three modification trials). Figure~\ref{fig:cogao4-volcano} summarizes this trajectory on a volcano plot obtained from Kulkarni et al.\cite{kulkarni2018understanding}.

\paragraph{Search summary:}
The agent's decision trace for \ce{CoGa2O4} demonstrates several key aspects of the autonomous reasoning loop:
\begin{enumerate}
    \item \textbf{Chemical hypothesis generation:} The agent drew on spinel crystal-chemistry knowledge to propose both compositional (Ga\(\to\)Al doping) and strain modifications as plausible tuning strategies.
    \item \textbf{Systematic isolation of variables:} After the combined doping-plus-strain attempt failed, the agent removed the dopant to isolate the strain effect, a standard experimental-design principle applied autonomously.
    \item \textbf{Feedback-driven sign reversal:} The quantitative AdsorbML feedback revealed that compressive strain strengthened adsorption on \ce{CoGa2O4}, contrary to the initial expectation. The agent reversed the strain direction in the next trial, demonstrating adaptive reasoning.
    
    \item \textbf{Convergence within few trials:} The agent reached an optimal condition within four evaluations.
    
    \item \textbf{Efficient modification selection:} Over three modification attempts the agent achieved a 2:1 success-to-failure ratio.
\end{enumerate}

This example underscores that Catalyst-Agent does not follow a pre-programmed hard-coded script, but instead generates, tests and revises modification hypotheses in real time, guided by physically grounded energy feedback.

\paragraph{Critical assessment of the agent's decision-making:}
Despite converging within four trials, the agent's verbal reasoning sometimes inverts thermodynamic language: the positive baseline $E_{\mathrm{ads}} = 1.1102$~eV indicates weak adsorption, yet the agent labeled it ``too-strong'' binding and aimed to ``soften'' it. When modifications raised $E_{\mathrm{ads}}$ further, the agent correctly identified these as moves in the wrong direction without reconciling this with its own weakening objective. Convergence was achieved through accurate numerical tracking of the descriptor rather than physically consistent reasoning. Moreover, the ``disentanglement'' in Trial~3 was a corrective step forced by the agent's own choice to separate the doping and strain effects employed in Trial~2, and the subsequent ``sign reversal``d of strain direction could be interpreted as exhaustive search over a binary parameter rather than a prediction grounded in electronic-structure reasoning about Co--O covalency or $d$-band shifts. There is trial and error with intelligent physical understanding at play here.

\subsubsection*{\ce{Tl3La} for \ce{CO2}RR}
\label{sec:si:tl3la-co2rr-adaptive-search}

The agent also demonstrated its ability by screening the \ce{Tl3La} surface (which is reported as LaTl3 in the CIF file, space group $Pm\bar{3}m$), on the (111) facet by checking that it met both of the \(\ast\)CO/\(\ast\)H pair criteria. The unstrained surface had already been found to be within the \(\ast\)CO adsorption window (\(E_{\text{ads}}(\ast\text{CO})=-0.6130\) eV), which meant it passed the stability test at -0.347 eV/atom. However, it failed the first criterion for hydrogen adsorption (\(E_{\text{ads}}(\ast\text{H})=+0.2369\) eV), since this was lower than the +0.33 eV limit for avoiding H-adsorbing surfaces that can lead to the competitive HER.

Therefore, the agent applied a minimal modification of 5\% in-plane compression. This preserved the CO binding energy at \(E_{\mathrm{ads}}(\ast\mathrm{CO}) =-0.6190\) eV, and moved the hydrogen adsorption energy from \(E_{\mathrm{ads}}(\ast\mathrm{H}) = +0.2369\)~eV to  \(E_{\mathrm{ads}}(\ast\mathrm{H}) = +0.3315\)~eV, so that this was now within the selected range. Therefore, Tl$_3$La was finally accepted as being CO\(_2\)RR selective after only two rounds, one of which was a run without any strain.

\subsection*{Closed-loop screening effectiveness across tasks}
\label{sec:results:success-metrics}

To measure how effectively Catalyst-Agent's screening process works, we use four performance metrics calculated across the three reaction-screening experiments, with each metric corresponding to a distinct aspect of autonomous screening. First, the success rate measures the fraction of tested materials that satisfy the target descriptors and therefore quantifies the overall coverage of the agent's search strategy. Second, the trials-to-success distribution measures how many evaluation attempts are required before a successful descriptor match is found: lower values indicate faster convergence and lower computational effort, whereas higher values indicate that materials require extended exploration. Third, the modification frequency measures the fraction of materials in each outcome category (successful or failed) that enters the structural modification loop. A higher modification frequency among successful materials means that descriptor-matching candidates often require modifications, while a higher modification frequency among failed materials indicates that the agent is spending additional effort on near-miss or difficult candidates that may remain outside the target window. Fourth, the number and distribution of successful and unsuccessful modifications among materials that entered the modification loop (Total Trials~$> 1$) quantify the productivity of those refinement attempts: more successful modifications indicate effective descriptor-directed tuning, whereas more unsuccessful modifications indicate exploratory or non-convergent search. A ``trial`` corresponds to a single adsorption-energy evaluation for a specific material-adsorbate configuration (material, surface facet, with or without applied modifications).

Let \(N_{\mathrm{tot}}\) be the total number of distinct materials evaluated for a given reaction screening task and let \(N_{\mathrm{succ}}\) be the number of materials for which at least one attempted condition satisfies the target descriptors. The success rate is:

\begin{equation}
R_{\mathrm{succ}} = \frac{N_{\mathrm{succ}}}{N_{\mathrm{tot}}}.
\end{equation}

A higher \(R_{\mathrm{succ}}\) implies that the candidate-generation and evaluation loop identifies a larger fraction of descriptor-satisfying materials in the sampled search space.

For each successful material \(i\), let \(t_i\) denote the number of trials required until the first successful condition is found (counting from the base, unmodified condition and including any subsequent facets and modification attempts). The average trials-to-success is then:

\begin{equation}
\overline{t}_{\mathrm{succ}} = \frac{1}{N_{\mathrm{succ}}}\sum_{i=1}^{N_{\mathrm{succ}}} t_i.
\end{equation}

A lower \(\overline{t}_{\mathrm{succ}}\) indicates that the agent satisfies the target descriptors with fewer adsorption-energy evaluations, while a higher value indicates a greater need for iterative exploration.

The modification frequency for a given outcome category (successful or failed materials) is the fraction of materials whose total trial count exceeds one:
\begin{equation}
f_{\mathrm{mod}} = \frac{N_{\mathrm{mod}}}{N_{\mathrm{cat}}},
\end{equation}
where \(N_{\mathrm{mod}}\) is the number of materials with Total Trials~$> 1$ in the category and \(N_{\mathrm{cat}}\) is the total number of materials in that category. This metric is interpreted separately for successful and failed categories: high \(f_{\mathrm{mod}}\) among successful materials indicates that structural refinement is an important route to finding hits, whereas high \(f_{\mathrm{mod}}\) among failed materials indicates persistent but ultimately unsuccessful rescue attempts. For materials that enter the modification loop, we further report the mean number of successful modifications and unsuccessful modifications per material, along with their standard deviations. The balance between these two counts indicates whether the modification loop is productive or mainly exploratory.

Together, \(R_{\mathrm{succ}}\) captures search coverage, \(\overline{t}_{\mathrm{succ}}\) captures convergence efficiency, \(f_{\mathrm{mod}}\) captures how often the agent chooses to refine candidates, and the per-material modification counts capture whether those refinements move materials toward or away from the descriptor targets.

\begin{table}[H]
\centering
\caption{Closed-loop screening outcomes across tasks. \(R_{\mathrm{succ}}\) is the fraction of tested materials that achieved at least one condition satisfying the task’s descriptor criteria. \(\overline{t}_{\mathrm{succ}}\) is the mean number of trials required to reach the first success, averaged over successful materials; \(\sigma_t\) is the corresponding standard deviation quantifying inter-material variability. \(N_{\mathrm{tot}}\) is the total number of distinct materials evaluated per task.}
\label{tab:success-rate-trials}
\large
\begin{tabular}{@{}lcccc@{}}
\toprule
\textbf{Task} & \(\mathbf{N_{\mathrm{tot}}}\) & \(\mathbf{R_{\mathrm{succ}}}\) & \(\mathbf{\overline{t}_{\mathrm{succ}}}\) & \(\boldsymbol{\sigma_t}\) \\
\midrule
ORR (via \(\ast\)OH descriptor) & 82 & 0.41 & 2.26 & 1.64 \\
NRR (via \(\ast\)N/\(\ast\)N\(_2\) descriptors) & 52 & 0.33 & 3.41 & 2.53 \\
CO\(_2\)RR (via \(\ast\)CO/\(\ast\)H descriptors) & 13 & 0.38 & 1.40 & 0.55 \\
\bottomrule
\end{tabular}
\end{table}

\begin{figure}[H]
\centering
\includegraphics[width=1.05\textwidth]{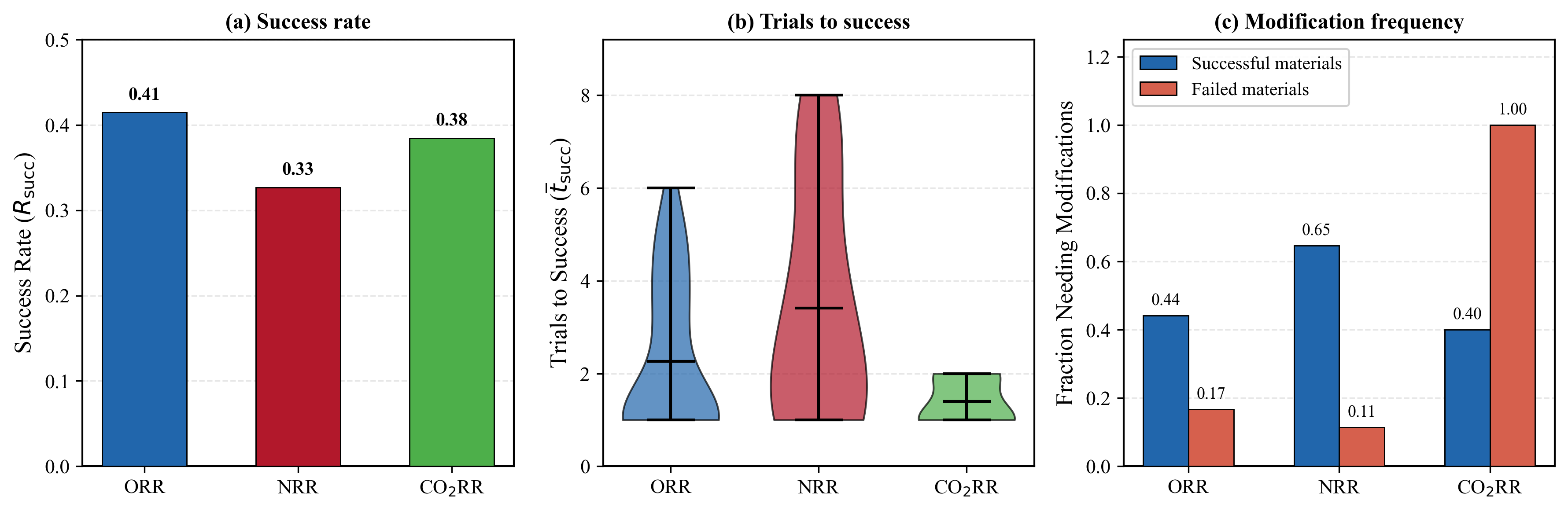}
\caption{Screening campaign overview across the three target reactions. (a)~Success rate \(R_{\mathrm{succ}}\). (b)~Distribution of trials to first success for successful materials: horizontal lines indicate mean and min--max range. (c)~Fraction of materials that required at least one structural modification (trial count ${>}\,1$), compared between successful and failed material categories.}
\label{fig:screening-overview}
\end{figure}

Multiple trends are evident from Table~\ref{tab:success-rate-trials} (per-material data is given in Section~S3). The success rates span a comparable range across the three case-study reactions (\(R_{\mathrm{succ}} \sim 0.33\)--0.41), which suggests that the agent's operational loop is able to identify promising candidates that meet the user-defined criteria across chemically different search spaces. This may appear low at first glance, but is higher than the results obtained by the LLM acting on its reasoning alone in the subsequent ablation study section.

CO\(_2\)RR shows the lowest \(\overline{t}_{\mathrm{succ}}\) (1.40\(\pm\)0.55), suggesting that successful candidates for CO\(_2\)RR are identified quickly with minimal iteration and good inter-material consistency, in contrast to the 100\% failure rate in the subsequent ablation study. But ORR and NRR show higher numbers of mean trials (2.26\(\pm\)1.64 and 3.41\(\pm\)2.53, respectively), indicating a greater need for multi-facet exploration and iterative structure modifications to land within their respective narrow target regions. The plot in Figure~\ref{fig:screening-overview}(b) shows that the trials-to-success distribution is right-skewed for ORR and NRR: many materials succeed on the first baseline evaluation, while a subset requires several rounds of strain or doping modifications. NRR displays the lowest success rate (0.33) among the three tasks, which is consistent with the stricter acceptance criterion that requires moderate \(\ast\)N binding and a favourable \(\Delta = E_{\mathrm{ads}}(\ast\mathrm{N}) - E_{\mathrm{ads}}(\ast\mathrm{N}_2) > 0\).

An overview of the screening campaign statistics is shown in Figure~\ref{fig:screening-overview}. For ORR, 44\% of successful materials required modifications compared to only 17\% of failures. For NRR, this contrast is even more pronounced: 65\% of successful materials needed modifying versus 11\% of failed materials. This shows that the agent invests more effort into promising materials. For CO\(_2\)RR, all of the faied materials (100\%) underwent multiple modification attempts, while only 40\% of successful materials required any modification. But many of the materials that failed are still quite promising, as their performance is close to the target window. They can be referred to in Section~S3 of the Supplementary Information. This asymmetry reflects the agent's persistent but ultimately futile effort to rescue candidates through modifications.


Overall, these metrics support two complementary takeaways: Catalyst-Agent has significant coverage across diverse electrocatalysis objectives (\(R_{\mathrm{succ}} \sim 0.33\)--0.41), and its iterative modification loop is efficiently utilized, converging in approximately 1-3.5 trials for successful cases on average. The low average number of trials indicates good initial candidate suggestion through the Catalyst Information Server. 

The agent appears to be particularly persistent when attempting to modify promising materials that ultimately fail to meet the desired criteria: many trials are conducted, particularly in the CO\(_2\)RR case. These results suggest that future screening campaigns may benefit from filtering out candidates based upon their proximity to the target descriptors prior to devoting time to modifying them. Notably, this is the approximate strategy that the agent employs through its logic for identifying promising near misses.

\subsection*{Ablation study}
\label{sec:si:internet-denied-ablation}

To determine the components of the Catalyst-Agent system which contribute most to its ability to screen materials successfully, we created four ablations of the full workflow, as shown in Table~\ref{tab:si-ablation-success-rates}. First, we removed web search from the candidate-discovery process while retaining the rest of the system's structure-retrieval and energy evaluation functionality. This condition is referred to as ``internet-denied'' in this section. Second, the agentic material selection method was replaced with random sampling from a list of Materials Project identifiers. Third, the agent was forced to accept a randomly selected modification for a given material.
Finally, we created a fourth ablation, called ``fully randomized``, in which both material selection and modification are random. In randomized settings, candidates are selected from a list containing approximately 160,000 unique Materials Project IDs. Once a candidate is selected, the modification type may be chosen randomly too. If a strain modification is selected, the strain value is randomly chosen between \(-0.08\) and \(+0.08\).

\begin{table*}[!htbp]
\centering
\caption{Ablation screening success rates for the original workflow and randomized baselines. Each entry reports the percentage of evaluated materials that achieved at least one accepted condition; counts are shown in parentheses when available.}
\label{tab:si-ablation-success-rates}
\begingroup
\setlength{\tabcolsep}{3pt}
\small
\renewcommand{\arraystretch}{0.9}
\begin{tabular}{@{}L{2.7cm}L{3.2cm}L{3.1cm}>{\centering\arraybackslash}p{1.7cm}>{\centering\arraybackslash}p{1.7cm}>{\centering\arraybackslash}p{1.8cm}@{}}
\toprule
\textbf{Condition} &
\textbf{Candidate source} &
\textbf{Modification policy} &
\textbf{ORR} &
\textbf{NRR} &
\textbf{CO\(_2\)RR} \\
\midrule
Original, fully agentic & Literature-informed and agent-selected & Feedback-directed by agent & 41\% & 33\% & 38\% \\
Internet-denied ablation & Agent-selected without web search & Feedback-directed by agent & 36.7\% (11/30) & 33.3\% (10/30) & 0\% (0/30) \\
Random selection only & Random Materials Project ID sampling & Feedback-directed by agent & 25.0\% (10/40) & 26.7\% (8/30) & 3.3\% (1/30) \\
Random modification only & Agent-selected & Randomly selected modification & 23.3\% (7/30) & 16.7\% (5/30) & 0\% (0/30) \\
Fully randomized & Random Materials Project ID sampling &  Randomly selected modification & 13.3\% (4/30) & 16.7\% (5/30) & 0\% (0/30) \\
\bottomrule
\end{tabular}
\endgroup
\end{table*}

Our results clearly demonstrate that the observed success rates are due to more than simply generating many AdsorbML evaluations. Instead, they depend upon combining chemically informed candidate selection with feedback-directed modification. As expected, the original fully agentic workflow produces the highest success rate for ORR and CO\(_2\)RR and continues to perform well relative to the ablated workflows for NRR.
Denying internet access only slightly impacts the agent's success rates for ORR and has no effect on the NRR case. The agent produced success rates of 36.7\% and 33.3\% for ORR and NRR, respectively, in the internet-denied condition, suggesting that the model retains useful internal chemical prior knowledge in its parameters relevant to these tasks. However, removing internet-enabled candidate selection severely impacts the agent's ability to find successful CO\(_2\)RR materials. In the internet-denied condition, zero out of 30 CO\(_2\)RR candidates succeeded, suggesting that the coupled \(\ast\)CO/\(\ast\)H descriptor criteria are especially sensitive to externally grounded candidate sourcing rather than internal-model selection alone based on parametric knowledge.

Randomizing material selection significantly impacts success rates for all three reactions.  Additionally, while random candidate selection is detrimental, randomizing the modification selection is even worse for ORR and NRR. With respect to modifying materials to improve their ability to achieve desired adsorption energies, randomizing modification selection drops the agent's success rates to 23.3\% and 16.7\% for ORR and NRR, respectively. These results demonstrate that the agent's modification choices are not random, instead, they contain useful information gained from the energy evaluation feedback. The fully randomized baseline performs worst of all, with success rates of 13.3\%, 16.7\%, and 0\% for ORR, NRR, and CO\(_2\)RR, respectively. These trends suggest that the best screening performance occurs when there is an agentic loop consisting of selecting a set of reasonable materials, evaluating adsorption energy descriptor properties for those materials, and then allowing the agent to decide targeted structural modifications to reduce the energetic gap identified during evaluation.

\section*{CO\(_2\)RR materials}
\label{sec:co2rr}

 Table~\ref{tab:co2rr-cathub-candidates} summarizes the five successful CO\(_2\)RR systems from the Catalyst-Agent run. These structures were obtained from Catalysis-Hub.org\cite{winther2019catalysis} entries reported by Jens S. Hummelshoj. No associated published manuscript was found for these entries, therefore, they are treated here as database-reported systems and novel screening hypotheses rather than literature-validated catalysts. The entries are viewed as candidates for further independent validation.

\begin{table*}[!htbp]
\centering
\caption{Successful CO\(_2\)RR candidates from the Catalysis-Hub.org portion of the screening campaign. The descriptor criteria are \(E_{\mathrm{ads}}(\ast\mathrm{CO}) \in [-0.82,-0.52]\) eV and \(E_{\mathrm{ads}}(\ast\mathrm{H}) \geq +0.33\) eV, with a stability screen of formation energy \(< -0.1\) eV/atom when reported. N/R denotes a value not reported in the run log.}
\label{tab:co2rr-cathub-candidates}
\small
\setlength{\tabcolsep}{4pt}
\renewcommand{\arraystretch}{1.25}
\resizebox{\textwidth}{!}{%
\begin{tabular}{@{}L{1.49cm}L{1.48cm}cL{2.55cm}cccL{4.75cm}@{}}
\toprule
\textbf{Screened material} & \textbf{CIF formula} & \textbf{Facet} & \textbf{Passing trial} & \makecell{\textbf{\(E_{\mathrm{ads}}\)} \\ \textbf{(\(\ast\)CO) [eV]}} (ML) & \makecell{\textbf{\(E_{\mathrm{ads}}\)} \\ \textbf{(\(\ast\)H) [eV]}} (ML) & \makecell{\textbf{Formation} \\ \textbf{energy [eV/atom]}} \\
\midrule
Pb$_3$Y & Pb$_3$Y & (111) & Pristine (run 0) & $-0.7384$ & $+0.3791$ & $-0.287$  \\
\addlinespace
Tl$_3$La & LaTl$_3$ & (111) & $+5\%$ compressive strain (run 1) & $-0.6190$ & $+0.3315$ & $-0.347$  \\
\addlinespace
Sn$_3$Y & Sn$_3$Y & (111) & Pristine (run 0) & $-0.6199$ & $+0.5129$ & $-0.495$ \\
\addlinespace
Sn$_3$Sc & ScSn$_3$ & (111) & Pristine (run 0) & $-0.7299$ & $+0.4646$ & $-0.335$  \\
\addlinespace
In$_3$Y & In$_3$Y & (111) & $+8.5\%$ compressive strain (run 1) & $-0.7737$ & $+0.3335$ & -0.429\\
\bottomrule
\end{tabular}%
}
\end{table*}

\clearpage 

\section*{DFT validation of selected candidates}
\label{sec:dft-validation}

To determine whether the screening decisions made using ML remain valid at higher accuracy and to show the robustness of the UMA relaxation, we performed single-point DFT validation on representative candidates from the NRR and CO\(_2\)RR campaigns. Three materials were selected for this analysis. The first was an NRR material that satisfied the descriptor criteria and was therefore classified as successful. The second was a CO\(_2\)RR candidate that satisfied some, but not all, of the descriptor criteria and was identified as a promising near-boundary candidate. The third was a CO\(_2\)RR candidate that satisfied all descriptor criteria and therefore showed strong promise. For each material, the same facets and adsorbates evaluated during Catalyst-Agent screening were reassessed with DFT, and the resulting adsorption energies were compared with the ML estimates as shown in Table \ref{tab:dft-ml-validation}.

\begin{table*}[!htbp]
\centering
\caption{DFT validation results with ML-predicted adsorption energies for comparison.}
\label{tab:dft-ml-validation}
\small
\setlength{\tabcolsep}{4pt}
\renewcommand{\arraystretch}{1.25}
\resizebox{\textwidth}{!}{%
\begin{tabular}{@{}llcllp{4.5cm}l@{}}
\toprule
\textbf{Material} & \textbf{Reaction} & \textbf{Facet} & \textbf{E($\ast A_1$) [eV]} & \textbf{E($\ast A_2$) [eV]} & \textbf{Criteria} & \textbf{Result} \\
\midrule
ReIr (mp-1219533) & NRR    & (110) & \makecell[l]{$\ast$N\\ DFT: $-0.611$\\ ML: $-0.556$}  & \makecell[l]{$\ast$N$_2$\\ DFT: $-0.645$\\ ML: $-0.704$} & $\ast$N $\in [-2.0, -0.5]$,\newline $\Delta = +0.034 > 0$ & PASS \\
\addlinespace
Sn$_3$Sc  & CO$_2$RR  & (111) & \makecell[l]{$\ast$CO\\ DFT: $-0.848$\\ ML: $-0.727$} & \makecell[l]{$\ast$H\\ DFT: $+0.494$\\ ML: $+0.466$}     & $\ast$CO outside $[-0.82, -0.52]$ (too strong);\newline $\ast$H $\geq 0.33$ \cmark & FAIL (almost pass) \\
\addlinespace
Sn$_3$Y   & CO$_2$RR  & (111) & \makecell[l]{$\ast$CO\\ DFT: $-0.673$\\ ML: $-0.621$} & \makecell[l]{$\ast$H\\ DFT: $+0.559$\\ ML: $+0.512$}     & $\ast$CO $\in [-0.82, -0.52]$ \cmark,\newline $\ast$H $\geq 0.33$ \cmark & PASS \\
\bottomrule
\end{tabular}%
}
\end{table*}

The DFT calculations preserve the qualitative screening outcome for ReIr in the NRR task and Sn$_3$Y in the CO\(_2\)RR task. ReIr satisfies the moderate $\ast$N binding window and maintains a positive $\Delta$ value, while Sn$_3$Y satisfies both the $\ast$CO and $\ast$H constraints for CO\(_2\)RR selectivity. Sn$_3$Sc remains a chemically promising near miss: DFT confirms unfavorable hydrogen adsorption relative to the HER suppression criterion, but the $\ast$CO adsorption energy is slightly more negative than the accepted CO\(_2\)RR window. These results support the use of Catalyst-Agent as a rapid hypothesis-generation and triage framework, while also emphasizing that final catalyst recommendations should be validated with higher-fidelity calculations and experiments when candidates lie close to descriptor boundaries. Further details about these calculations can be found in the Methods section.








\section*{Conclusions}
\label{sec:conclusion}

We present Catalyst-Agent, an autonomous, tool-grounded LLM agent that operationalizes catalyst screening workflows as closed-loop processing campaigns. The agent is able to translate scientific natural language commands and objectives into reproducible sequences of actions: finding candidate materials, slab construction, adsorption energy calculation, constraint evaluation, and iterative refinement via simple structural modifications. Across the ORR, NRR, and CO$_2$RR-motivated target descriptors, the agent achieved a reasonable success rate across all catalyst candidates and converged to success within \(\sim 1\)-3.5 trials on average when success was possible, illustrating that autonomous processing and iteration can reduce the human researcher's time and effort in computational catalysis exploration. Furthermore, the fact that the agent identified Sn$_3$Sc, Sn$_3$Y, Tl$_3$La, Pb$_3$Y and In$_3$Y as potential candidates for further validation, materials not previously reported in the literature (as shown in Table~\ref{tab:co2rr-cathub-candidates}), demonstrates the ability of LLM-based agents to provide catalyst screening opportunities. The ablation study further shows that this performance arises from chemically informed candidate selection and feedback-directed modification, rather than random trial and error actions, with literature-informed sourcing being particularly important for the more restrictive CO$_2$RR descriptor criteria. Future work would involve additional modification strategies, fine-tuning MLIPs on specific chemistries, and systematically evaluating LLM hallucinations.

The results provide a proof of concept for LLM-powered agents integrated with advanced computational scientific tools that can execute end-to-end catalyst screening with minimal human intervention, within a tool-grounded workflow. For a practical scientific coding scenario, the modular MCP server architecture isolates conflicting software dependencies: improving maintainability, reusability, and alterability, while the generous free tiers and rate limits of current LLM APIs keep short screening campaigns accessible at negligible cost, though larger-scale campaigns will incur higher API expenses (token usage and costs are detailed in Section~S5). This work creates a practical pathway toward rapid, scalable, and reproducible computational screening with tool-grounded adaptive search. Human effort can now be shifted from scripting to defining objectives, high-level orchestration, and pattern recognition.

\section*{Technology Use Disclosure}
ChatGPT and Claude were utilized for grammatical refinement and typographical corrections. Subparts of Figures 1 and 2 were also derived from these tools. The authors have carefully reviewed and verified all sections of the manuscript for accuracy and integrity.

\section*{Data and Software Availability Statement}
For public benefit and further research, the code, raw prompt and output files are available at the following link: \url{https://github.com/BaratiLab/Catalyst_agent}.

\begin{acknowledgement}

The authors thank Omid Barati, Abraham George, Yayati Jadhav, Radheesh Sharma Meda, and Yue Su for insightful discussions and helpful suggestions.

\end{acknowledgement}

\section*{Supporting Information}
Supporting Information is available and includes the following sections: Example user prompts, Example MCP tool calls, Agent outputs by reaction, Illustrative example of agentic search, DFT calculation settings, An ablation study, UMA performance metrics, and token consumption statistics.

\bibliography{references}

\clearpage
\pagenumbering{arabic}
\renewcommand*{\thepage}{S\arabic{page}}
\setcounter{subsection}{0}
\setcounter{figure}{0}
\setcounter{table}{0}
\renewcommand{\thesubsection}{S\arabic{subsection}}
\renewcommand{\thesubsubsection}{S\arabic{subsection}.\arabic{subsubsection}}
\renewcommand{\thefigure}{S\arabic{figure}}
\renewcommand{\thetable}{S\arabic{table}}

\section*{Supplemental Information}
\addcontentsline{toc}{section}{Supplemental Information}
\subsection*{Contents}
\begin{enumerate}
    \item \hyperref[sec:si:example-prompt]{Example User Prompt: ORR Screening with Iterative Modification}
    \item \hyperref[sec:si:example-calls]{Example MCP Tool Calls: \ce{CoAl2O4} (mp-36447)}
    \item \hyperref[sec:si:raw-output]{Agent Output by Reaction}
    \item \hyperref[sec:results:uma-predictive-power]{Predictive performance of UMA for catalysis: OC20 benchmarks}
    \item \hyperref[sec:si:token_cost]{Token consumption and costs}
\end{enumerate}
\subsection{Example User Prompt: ORR Screening with Iterative Modification}
\label{sec:si:example-prompt}

The following is a representative user prompt submitted to Catalyst-Agent for an end-to-end ORR screening campaign. This prompt demonstrates how a researcher specifies the target reaction, adsorbate, acceptance criteria, tool-server routing and iterative modification strategy in natural language. The agent parses these instructions and autonomously plans the sequence of MCP tool calls required to execute the screening workflow.

\begin{lstlisting}[style=bluecode]
Use the perplexity mcp server to get a list of good candidates
for the oxygen reduction reaction. Use the structure retrieval
mcp server to get the pymatgen structures. Run the energy
evaluation server using the *OH adsorbate. The ideal energy
should be between 0.9-1.1 eV. You may apply modifications using
the structure modification mcp server after trying a run with no modifications. Please search and run one material at a time. You may also search for materials that are not suggested, but those which you deem promising. Please document all observations and your reasoning behind them in a local text document in real-time.
\end{lstlisting}

\noindent This prompt encodes several operational directives that the agent decomposes into discrete planning steps:

\begin{enumerate}
    \item \textbf{Candidate sourcing:} The Catalyst Information Server (backed by Perplexity Deep Research) is queried first to obtain a literature-informed list of promising oxide compositions for ORR.
    \item \textbf{Structure retrieval:} For each candidate, the Structure Retrieval Server queries the OPTIMADE API to obtain serialized pymatgen structure objects from the Materials Project and OQMD.
    \item \textbf{Baseline evaluation:} The Energy Evaluation Server computes $\ast$OH adsorption energies on multiple Miller-index facets using the AdsorbML--UMA pipeline, with the target descriptor window set to 0.9--1.1~eV.
    \item \textbf{Conditional modification:} Only after an unmodified baseline run does the agent invoke the Structure Modification Server to apply strain or top-layer substitution, guided by the energetic feedback from the baseline.
    \item \textbf{Sequential processing:} The instruction to ``search and run one material at a time'' directs the agent to complete the full evaluate--modify--re-evaluate loop for each candidate before proceeding to the next, ensuring that context-window usage remains manageable and that per-material decision traces are cleanly separated.
\end{enumerate}

\subsection{Example MCP Tool Calls: \ce{CoAl2O4} (mp-36447)}
\label{sec:si:example-calls}

This section presents the complete sequence of MCP tool calls executed by Catalyst-Agent during the screening of \ce{CoAl2O4} (Materials Project identifier mp-36447, space group $Fd\bar{3}m$) for the oxygen reduction reaction using the $\ast$OH adsorbate. Each call is shown with its tool name, input parameters (JSON), and the returned output (JSON). The outer MCP protocol envelope (\texttt{"result": "..."}) is unwrapped for clarity; only the parsed payload is shown. Calls~1--5 correspond to structure retrieval and unstrained adsorption energy evaluations across five Miller-index facets. Calls~6--9 repeat the adsorption analysis on the same facets under 2\% compressive in-plane strain. All energies are in~eV.

\subsubsection{Call 1: Structure Search via OPTIMADE}

\noindent\textbf{Tool:} \texttt{optimade\_structure\_search}

\noindent\textbf{Input:}
\begin{lstlisting}[style=json]
{
  "elements": ["Co", "Al", "O"],
  "nelements": 3
}
\end{lstlisting}

\noindent\textbf{Output} (truncated; 49 structures returned across Materials Project and OQMD):
\begin{lstlisting}[style=json]
{
  "message": "Found 49 structures matching criteria",
  "results_summary": [
    {
      "provider": "mp",
      "identifier": "mp-1435808",
      "formula": "AlCoO3",
      "spacegroup": "Pm-3m"
    },
    {
      "provider": "mp",
      "identifier": "mp-1391681",
      "formula": "Al(CoO2)2",
      "spacegroup": "C2/c"
    },
    {
      "provider": "mp",
      "identifier": "mp-36447",
      "formula": "Al2CoO4",
      "spacegroup": "Fd-3m"
    },
    {
      "provider": "oqmd",
      "identifier": 4864833,
      "formula": "Al2CoO4",
      "spacegroup": "Fd-3m"
    }
  ],
  "files_saved": [
    "results.json",
    "results_short.json"
  ]
}
\end{lstlisting}

\noindent The full output includes 28 entries from the Materials Project and 21 from OQMD, spanning formulas \ce{AlCoO3}, \ce{Al(CoO2)2}, \ce{Al2CoO4}, \ce{Al2Co2O7}, \ce{Al4(CoO4)3}, \ce{Al5(CoO4)3}, \ce{Al(CoO2)3}, and \ce{Al2Co2O5}. The agent selected mp-36447 (\ce{Al2CoO4}, $Fd\bar{3}m$) for subsequent adsorption analysis.

\subsubsection{Calls 2--5: $\ast$OH Adsorption Energy (Unstrained)}

The following four calls evaluate $\ast$OH adsorption energies on mp-36447 across five Miller-index facets without surface modifications. Call~2 evaluates two facets in a single invocation; Calls~3--5 each evaluate one facet.

\paragraph{Call 2: (001) and (100) facets.}

\noindent\textbf{Tool:} \texttt{adsorbml\_evaluate}

\noindent\textbf{Input:}
\begin{lstlisting}[style=json]
{
  "provider": "mp",
  "identifier": "mp-36447",
  "adsorbate": "*OH",
  "hkl1": "[0,0,1]",
  "hkl2": "[1,0,0]"
}
\end{lstlisting}

\noindent\textbf{Output:}
\begin{lstlisting}[style=json]
{
  "provider": "mp",
  "identifier": "mp-36447",
  "adsorbate": "*OH",
  "cif_path": null,
  "doping": null,
  "strain": null,
  "results_by_hkl": {
    "(0,0,1)": {
      "provider": "mp",
      "identifier": "mp-36447",
      "formula": "Al2CoO4",
      "spacegroup": "Fd-3m",
      "spacegroup_number": null,
      "adsorbate": "*OH",
      "hkl": [0, 0, 1],
      "cif_path": null,
      "modifications_applied": null,
      "analysis_summary": {
        "total_configurations": 30,
        "valid_configurations": 28,
        "anomalies_detected": 3
      },
      "minimum_energy_results": {
        "config_index": 0,
        "adsorption_energy": 1.12311411448532,
        "slab_energy": -722.0156515494343,
        "gas_reactant_energy": -10.681,
        "adslab_energy": -731.573537434949
      }
    },
    "(1,0,0)": {
      "provider": "mp",
      "identifier": "mp-36447",
      "formula": "Al2CoO4",
      "spacegroup": "Fd-3m",
      "spacegroup_number": null,
      "adsorbate": "*OH",
      "hkl": [1, 0, 0],
      "cif_path": null,
      "modifications_applied": null,
      "analysis_summary": {
        "total_configurations": 30,
        "valid_configurations": 26,
        "anomalies_detected": 4
      },
      "minimum_energy_results": {
        "config_index": 0,
        "adsorption_energy": 1.1236939484699224,
        "slab_energy": -722.0156973258015,
        "gas_reactant_energy": -10.681,
        "adslab_energy": -731.5730033773316
      }
    }
  }
}
\end{lstlisting}

\paragraph{Call 3: (110) facet.}

\noindent\textbf{Tool:} \texttt{adsorbml\_evaluate}

\noindent\textbf{Input:}
\begin{lstlisting}[style=json]
{
  "provider": "mp",
  "identifier": "mp-36447",
  "adsorbate": "*OH",
  "hkl1": "[1,1,0]"
}
\end{lstlisting}

\noindent\textbf{Output:}
\begin{lstlisting}[style=json]
{
  "provider": "mp",
  "identifier": "mp-36447",
  "adsorbate": "*OH",
  "cif_path": null,
  "doping": null,
  "strain": null,
  "results_by_hkl": {
    "(1,1,0)": {
      "provider": "mp",
      "identifier": "mp-36447",
      "formula": "Al2CoO4",
      "spacegroup": "Fd-3m",
      "spacegroup_number": null,
      "adsorbate": "*OH",
      "hkl": [1, 1, 0],
      "cif_path": null,
      "modifications_applied": null,
      "analysis_summary": {
        "total_configurations": 30,
        "valid_configurations": 20,
        "anomalies_detected": 10
      },
      "minimum_energy_results": {
        "config_index": 0,
        "adsorption_energy": -1.3793883269203615,
        "slab_energy": -722.2220266715051,
        "gas_reactant_energy": -10.681,
        "adslab_energy": -734.2824149984255
      }
    }
  }
}
\end{lstlisting}

\paragraph{Call 4: (111) facet.}

\noindent\textbf{Tool:} \texttt{adsorbml\_evaluate}

\noindent\textbf{Input:}
\begin{lstlisting}[style=json]
{
  "provider": "mp",
  "identifier": "mp-36447",
  "adsorbate": "*OH",
  "hkl1": "[1,1,1]"
}
\end{lstlisting}

\noindent\textbf{Output:}
\begin{lstlisting}[style=json]
{
  "provider": "mp",
  "identifier": "mp-36447",
  "adsorbate": "*OH",
  "cif_path": null,
  "doping": null,
  "strain": null,
  "results_by_hkl": {
    "(1,1,1)": {
      "provider": "mp",
      "identifier": "mp-36447",
      "formula": "Al2CoO4",
      "spacegroup": "Fd-3m",
      "spacegroup_number": null,
      "adsorbate": "*OH",
      "hkl": [1, 1, 1],
      "cif_path": null,
      "modifications_applied": null,
      "analysis_summary": {
        "total_configurations": 30,
        "valid_configurations": 29,
        "anomalies_detected": 1
      },
      "minimum_energy_results": {
        "config_index": 0,
        "adsorption_energy": -4.679040426530305,
        "slab_energy": -1058.8612306200491,
        "gas_reactant_energy": -10.681,
        "adslab_energy": -1074.2212710465794
      }
    }
  }
}
\end{lstlisting}

\paragraph{Call 5: (210) facet.}

\noindent\textbf{Tool:} \texttt{adsorbml\_evaluate}

\noindent\textbf{Input:}
\begin{lstlisting}[style=json]
{
  "provider": "mp",
  "identifier": "mp-36447",
  "adsorbate": "*OH",
  "hkl1": "[2,1,0]"
}
\end{lstlisting}

\noindent\textbf{Output:}
\begin{lstlisting}[style=json]
{
  "provider": "mp",
  "identifier": "mp-36447",
  "adsorbate": "*OH",
  "cif_path": null,
  "doping": null,
  "strain": null,
  "results_by_hkl": {
    "(2,1,0)": {
      "provider": "mp",
      "identifier": "mp-36447",
      "formula": "Al2CoO4",
      "spacegroup": "Fd-3m",
      "spacegroup_number": null,
      "adsorbate": "*OH",
      "hkl": [2, 1, 0],
      "cif_path": null,
      "modifications_applied": null,
      "analysis_summary": {
        "total_configurations": 30,
        "valid_configurations": 26,
        "anomalies_detected": 4
      },
      "minimum_energy_results": {
        "config_index": 0,
        "adsorption_energy": -0.5089964812168954,
        "slab_energy": -542.2962311547523,
        "gas_reactant_energy": -10.681,
        "adslab_energy": -553.4862276359692
      }
    }
  }
}
\end{lstlisting}

\subsubsection{Calls 6--9: $\ast$OH Adsorption Energy (2\% Compressive Strain)}

The following four calls repeat the adsorption analysis on the same facets with 2\% compressive in-plane strain applied (\(\varepsilon=0.02\); the $a$ and $b$ lattice vectors are scaled by $1-\varepsilon$). The \texttt{modifications\_applied} field in each facet result now records the applied strain.

\paragraph{Call 6: (001) and (100) facets, 2\% strain.}

\noindent\textbf{Tool:} \texttt{adsorbml\_evaluate}

\noindent\textbf{Input:}
\begin{lstlisting}[style=json]
{
  "provider": "mp",
  "identifier": "mp-36447",
  "adsorbate": "*OH",
  "strain": "0.02",
  "hkl1": "[0,0,1]",
  "hkl2": "[1,0,0]"
}
\end{lstlisting}

\noindent\textbf{Output:}
\begin{lstlisting}[style=json]
{
  "provider": "mp",
  "identifier": "mp-36447",
  "adsorbate": "*OH",
  "cif_path": null,
  "doping": null,
  "strain": 0.02,
  "results_by_hkl": {
    "(0,0,1)": {
      "provider": "mp",
      "identifier": "mp-36447",
      "formula": "Al2CoO4",
      "spacegroup": "Fd-3m",
      "spacegroup_number": null,
      "adsorbate": "*OH",
      "hkl": [0, 0, 1],
      "cif_path": null,
      "modifications_applied": {
        "strain": {
          "value": 0.02,
          "percentage": 2.0,
          "type": "compressive"
        }
      },
      "analysis_summary": {
        "total_configurations": 30,
        "valid_configurations": 24,
        "anomalies_detected": 6
      },
      "minimum_energy_results": {
        "config_index": 0,
        "adsorption_energy": 1.0919098908526212,
        "slab_energy": -720.5685842886923,
        "gas_reactant_energy": -10.681,
        "adslab_energy": -730.1576743978396
      }
    },
    "(1,0,0)": {
      "provider": "mp",
      "identifier": "mp-36447",
      "formula": "Al2CoO4",
      "spacegroup": "Fd-3m",
      "spacegroup_number": null,
      "adsorbate": "*OH",
      "hkl": [1, 0, 0],
      "cif_path": null,
      "modifications_applied": {
        "strain": {
          "value": 0.02,
          "percentage": 2.0,
          "type": "compressive"
        }
      },
      "analysis_summary": {
        "total_configurations": 30,
        "valid_configurations": 29,
        "anomalies_detected": 1
      },
      "minimum_energy_results": {
        "config_index": 0,
        "adsorption_energy": 1.0919251496413427,
        "slab_energy": -720.5685842886919,
        "gas_reactant_energy": -10.681,
        "adslab_energy": -730.1576591390506
      }
    }
  }
}
\end{lstlisting}

\paragraph{Call 7: (110) facet, 2\% strain.}

\noindent\textbf{Tool:} \texttt{adsorbml\_evaluate}

\noindent\textbf{Input:}
\begin{lstlisting}[style=json]
{
  "provider": "mp",
  "identifier": "mp-36447",
  "adsorbate": "*OH",
  "strain": "0.02",
  "hkl1": "[1,1,0]"
}
\end{lstlisting}

\noindent\textbf{Output:}
\begin{lstlisting}[style=json]
{
  "provider": "mp",
  "identifier": "mp-36447",
  "adsorbate": "*OH",
  "cif_path": null,
  "doping": null,
  "strain": 0.02,
  "results_by_hkl": {
    "(1,1,0)": {
      "provider": "mp",
      "identifier": "mp-36447",
      "formula": "Al2CoO4",
      "spacegroup": "Fd-3m",
      "spacegroup_number": null,
      "adsorbate": "*OH",
      "hkl": [1, 1, 0],
      "cif_path": null,
      "modifications_applied": {
        "strain": {
          "value": 0.02,
          "percentage": 2.0,
          "type": "compressive"
        }
      },
      "analysis_summary": {
        "total_configurations": 30,
        "valid_configurations": 27,
        "anomalies_detected": 3
      },
      "minimum_energy_results": {
        "config_index": 0,
        "adsorption_energy": -1.4725890105157031,
        "slab_energy": -719.4819296256057,
        "gas_reactant_energy": -10.681,
        "adslab_energy": -731.6355186361214
      }
    }
  }
}
\end{lstlisting}

\paragraph{Call 8: (111) facet, 2\% strain.}

\noindent\textbf{Tool:} \texttt{adsorbml\_evaluate}

\noindent\textbf{Input:}
\begin{lstlisting}[style=json]
{
  "provider": "mp",
  "identifier": "mp-36447",
  "adsorbate": "*OH",
  "strain": "0.02",
  "hkl1": "[1,1,1]"
}
\end{lstlisting}

\noindent\textbf{Output:}
\begin{lstlisting}[style=json]
{
  "provider": "mp",
  "identifier": "mp-36447",
  "adsorbate": "*OH",
  "cif_path": null,
  "doping": null,
  "strain": 0.02,
  "results_by_hkl": {
    "(1,1,1)": {
      "provider": "mp",
      "identifier": "mp-36447",
      "formula": "Al2CoO4",
      "spacegroup": "Fd-3m",
      "spacegroup_number": null,
      "adsorbate": "*OH",
      "hkl": [1, 1, 1],
      "cif_path": null,
      "modifications_applied": {
        "strain": {
          "value": 0.02,
          "percentage": 2.0,
          "type": "compressive"
        }
      },
      "analysis_summary": {
        "total_configurations": 30,
        "valid_configurations": 30,
        "anomalies_detected": 0
      },
      "minimum_energy_results": {
        "config_index": 0,
        "adsorption_energy": -3.987527365007095,
        "slab_energy": -1055.4064118944632,
        "gas_reactant_energy": -10.681,
        "adslab_energy": -1070.0749392594703
      }
    }
  }
}
\end{lstlisting}

\paragraph{Call 9: (210) facet, 2\% strain.}

\noindent\textbf{Tool:} \texttt{adsorbml\_evaluate}

\noindent\textbf{Input:}
\begin{lstlisting}[style=json]
{
  "provider": "mp",
  "identifier": "mp-36447",
  "adsorbate": "*OH",
  "strain": "0.02",
  "hkl1": "[2,1,0]"
}
\end{lstlisting}

\noindent\textbf{Output:}
\begin{lstlisting}[style=json]
{
  "provider": "mp",
  "identifier": "mp-36447",
  "adsorbate": "*OH",
  "cif_path": null,
  "doping": null,
  "strain": 0.02,
  "results_by_hkl": {
    "(2,1,0)": {
      "provider": "mp",
      "identifier": "mp-36447",
      "formula": "Al2CoO4",
      "spacegroup": "Fd-3m",
      "spacegroup_number": null,
      "adsorbate": "*OH",
      "hkl": [2, 1, 0],
      "cif_path": null,
      "modifications_applied": {
        "strain": {
          "value": 0.02,
          "percentage": 2.0,
          "type": "compressive"
        }
      },
      "analysis_summary": {
        "total_configurations": 30,
        "valid_configurations": 24,
        "anomalies_detected": 6
      },
      "minimum_energy_results": {
        "config_index": 0,
        "adsorption_energy": -0.6058897917651347,
        "slab_energy": -540.8192414086575,
        "gas_reactant_energy": -10.681,
        "adslab_energy": -552.1061312004226
      }
    }
  }
}
\end{lstlisting}

\newpage
\subsection{Agent Output by Reaction}
\label{sec:si:raw-output}

Considering the large output sizes, the repository at \url{https://github.com/BaratiLab/Catalyst_agent} has:

\begin{enumerate}

    \item The material suggestions by the Catalyst Information Server at ORR\_materials.md, NRR\_materials.md
    
    \item The energy evaluation results files, named as ORR.txt, CO2RR.txt and NRR.txt respectively.

\end{enumerate}

\clearpage

\subsection{Predictive performance of UMA for catalysis: OC20 benchmarks}
\label{sec:results:uma-predictive-power}

The feasibility of closed-loop catalytic screening depends on the accuracy and the fidelity of the backbone MLIP's predictions of adsorption energies and forces across varied adsorbate-surface systems. UMA models\cite{wood2025family} provide a solid foundation for this purpose. These models achieve low total-energy and force errors on the OC20 dataset\cite{chanussot2021open}, which is a large-scale adsorption dataset based on DFT relaxations. They also improve adsorption energy prediction quality relative to prior baselines. Table~\ref{tab:catalysis_val_test_evals} reports OC20  validation errors for total energy (in-distribution and out-of-distribution splits) and OC20 test errors for adsorption energy, where adsorption energy is computed as a difference of two total-energy calculations (clean slab and adsorbate-slab). This is a formulation that empirically improves the accuracy of the adsorption-energy predictions over prior models.

Beyond pointwise energy and force MAE predictions, catalytic screening requires the determination of the global minimum energy across many adsorbate candidate placement sites. The AdsorbML framework\cite{lan2023adsorbml} enables the finding of this global minimum. A practical limitation of the original benchmark is the necessity of a downstream DFT evaluation of the ML-predicted global minima energy structure. To improve accessibility for those without the high-throughput DFT infrastructure, the ML-only variant replaces the DFT re-evaluation and is considered successful if the predicted energy is within 0.1 eV of the DFT minimum. This framework is well-aligned with Catalyst-Agent's deployment setting, where the agent must repeatedly generate and relax many adsorbate-catalyst geometries and then proceed with further decisions based on the minimum-energy predictions. Taken together, the OC20 evaluation metrics and AdsorbML pipeline motivate UMA as an effective MLIP backbone model for autonomous catalyst screening.

\begin{table*}[ht!]
\centering
\caption{UMA Catalysis validation and test results on OC20 metrics. All energies are in meV and forces are in meV/\AA.}
\label{tab:catalysis_val_test_evals}
\vspace{6pt}
\resizebox{\textwidth}{!}{
\begin{tabular}{L{3.0cm}cccccccccc}
\toprule
\multirow{3}{*}{\textbf{Model}} &
\multicolumn{6}{c}{\textit{Val (Total Energy)}} &
\multicolumn{4}{c}{\textit{Test (Ads. Energy)}} \\
\cmidrule(lr){2-7}\cmidrule(lr){8-11}
&
\multicolumn{3}{c}{ID} &
\multicolumn{3}{c}{OOD-Both} &
\multicolumn{2}{c}{ID} &
\multicolumn{2}{c}{OOD-Both} \\
\cmidrule(lr){2-4}\cmidrule(lr){5-7}\cmidrule(lr){8-9}\cmidrule(lr){10-11}
&
Energy & Forces & Force Cosine &
Energy & Forces & Force Cosine &
Energy & Force &
Energy & Force \\
\midrule
\multicolumn{11}{l}{\textbf{UMA}} \\
UMA-S & 63.6 & 24.1 & 0.63 & 107.0 & 29.2 & 0.65 & 52.1 & 24.3 & 70.2 & 30.9 \\
UMA-M & 43.1 & 15.8 & 0.73 & 70.0  & 19.2 & 0.75 & 33.4 & 16.0 & 46.5 & 21.0 \\
UMA-L & 32.6 & 12.0 & 0.77 & 49.8  & 14.5 & 0.79 & 32.4 & 12.2 & 43.5 & 15.9 \\
\midrule
\multicolumn{11}{l}{\textbf{Literature}} \\
eqV2-OC20~\cite{liao2023equiformerv2} &  --  &  --  & --  &  --  & --   &  -- & 149.1 & 11.63 & 306.5 & 15.74 \\
GemNet-OC20~\cite{gasteiger2022gemnet} &  --  & --   &  -- &  --  &  --  & --  & 163.5 & 16.33 & 343.3 & 23.11 \\
\bottomrule
\end{tabular}
}
\end{table*}

\clearpage

\subsection{Token consumption and costs}
\label{sec:si:token_cost}
\begin{table}[ht]
\centering
\fontsize{9.5pt}{9pt}\selectfont
\begin{tabular}{llrrr}
\toprule
Date & Model & Input & Output & Cost (USD) \\
\midrule
Dec 11, 2025 & gpt-5.2 & 1,916,125 & 109,717 & \$10.98 \\
Dec 12, 2025 & gpt-5.2 & 29,192 & 5,594 & \$0.14 \\
Dec 13, 2025 & gpt-5.2 & 1,394,281 & 233,962 & \$12.48 \\
Dec 14, 2025 & gpt-5.2 & 364,306 & 36,127 & \$1.68 \\
Dec 15, 2025 & gpt-5.2 & 593,190 & 58,199 & \$2.32 \\
Dec 16, 2025 & gpt-5.2 & 457,457 & 140,810 & \$2.76 \\
Dec 17, 2025 & gpt-5.2 & 183,519 & 1,542 & \$0.27 \\
Dec 18, 2025 & gpt-5.2 & 43,554 & 17,653 & \$0.42 \\
Dec 19, 2025 & gpt-5.2 & 337,487 & 34,899 & \$2.06 \\
Dec 20, 2025 & gpt-5.2 & 138,698 & 21,073 & \$0.62 \\
Dec 21, 2025 & gpt-5.2 & 1,180,587 & 250,224 & \$6.69 \\
Dec 22, 2025 & gpt-5.2 & 1,774,997 & 219,064 & \$11.43 \\
Dec 23, 2025 & gpt-5.2 & 186,313 & 34,148 & \$1.02 \\
Dec 24, 2025 & gpt-5.2 & 869,371 & 95,497 & \$5.65 \\
Dec 25, 2025 & gpt-5.2 & 645,791 & 125,096 & \$2.74 \\
Jan 07, 2026 & gpt-5.2 & 168,965 & 16,754 & \$0.63 \\
Jan 12, 2026 & gpt-5.2 & 571,119 & 89,322 & \$3.56 \\
Jan 13, 2026 & gpt-5.2 & 257,068 & 48,103 & \$1.46 \\
Feb 05, 2026 & gpt-5.2 & 107,268 & 50,044 & \$1.11 \\
May 02, 2026 & gpt-5.2 & 101,826 & 12,294 & \$0.53 \\
May 03, 2026 & gpt-5.2 & 816,147 & 82,345 & \$4.50 \\
May 04, 2026 & gpt-5.2 & 1,062,800 & 162,130 & \$7.66 \\
May 05, 2026 & gpt-5.2 & 876,171 & 82,018 & \$4.46 \\
May 06, 2026 & gpt-5.2 & 603,269 & 46,100 & \$4.09 \\
May 07, 2026 & gpt-5.2 & 965,946 & 122,786 & \$8.22 \\
May 08, 2026 & gpt-5.2 & 355,518 & 44,657 & \$2.20 \\
Jun 11, 2026 & gpt-5.2 & 44,778 & 3,177 & \$0.20 \\
Jun 12, 2026 & gpt-5.2 & 577,913 & 33,479 & \$2.91 \\
Jun 13, 2026 & gpt-5.2 & 567,896 & 62,030 & \$3.96 \\
Jun 14, 2026 & gpt-5.2 & 846,205 & 81,078 & \$7.21 \\
Jun 15, 2026 & gpt-5.2 & 950,716 & 75,222 & \$8.82 \\
Jun 16, 2026 & gpt-5.2 & 830,238 & 90,839 & \$7.33 \\
Jun 17, 2026 & gpt-5.2 & 992,163 & 84,178 & \$8.58 \\
Jun 18, 2026 & gpt-5.2 & 1,742,688 & 119,023 & \$15.42 \\
Jun 19, 2026 & gpt-5.2 & 1,781,932 & 94,401 & \$14.68 \\
Jun 20, 2026 & gpt-5.2 & 1,048,024 & 70,062 & \$9.77 \\
Jun 21, 2026 & gpt-5.2 & 1,069,904 & 76,454 & \$9.80 \\
Jun 22, 2026 & gpt-5.2 & 88,682 & 4,727 & \$0.35 \\
Jun 24, 2026 & gpt-5.2 & 85,315 & 5,419 & \$0.39 \\
Jun 25, 2026 & gpt-5.2 & 1,379,382 & 99,599 & \$8.12 \\
Jun 26, 2026 & gpt-5.2 & 888,344 & 72,077 & \$4.91 \\
Jun 27, 2026 & gpt-5.2 & 691,029 & 58,653 & \$5.16 \\
\midrule
\textbf{Total} &  & \textbf{29,586,174} & \textbf{3,170,576} & \textbf{\$207.29} \\
\bottomrule
\end{tabular}
\caption{Token usage: Both inputs and outputs with costs}
\end{table}

\clearpage

\end{document}